\documentclass[10pt,twocolumn,letterpaper]{article}

\usepackage[accsupp]{axessibility}
\usepackage{iccv}              
\usepackage{todonotes}
\usepackage{graphicx}
\usepackage{subcaption}
\usepackage{multirow}

\definecolor{iccvblue}{rgb}{0.21,0.49,0.74}
\usepackage[pagebackref,breaklinks,colorlinks,allcolors=iccvblue]{hyperref}

\def\paperID{13331} 
\def\confName{ICCV}
\def\confYear{2025}

\title{EmbodiedSplat: Personalized Real-to-Sim-to-Real Navigation with Gaussian Splats from a Mobile Device}

\author{
Gunjan Chhablani$^1$\thanks{Work done as a student at Georgia Tech, currently at Waymo.}, Xiaomeng Ye$^2$, Muhammad Zubair Irshad$^3$, Zsolt Kira$^4$\\
$^{1,2,4}$Georgia Tech, $^3$Toyota Research Institute\\
{\tt\small \{$^1$chhablani.gunjan, $^3$muhammadzubairirshad\}@gmail.com, \{$^2$xye87, $^4$zkira\}@gatech.edu}
}

\newcommand{\msr}{\texttt{MuSHRoom}\xspace
}
\newcommand{\cpt}{\texttt{Captured}\xspace}
\newcommand{\csr}{\texttt{classroom}\xspace
}
\newcommand{\lng}{\texttt{lounge}\xspace
}
\newcommand{\pid}{\texttt{conf\_a}\xspace
}
\newcommand{\cab}{\texttt{conf\_b}\xspace
}
\newcommand{\poly}{\textsc{Polycam}\xspace}

\newcommand{\dn}{\textsc{DN}\xspace}

\begin{document}
\maketitle
\begin{abstract}
\looseness=-1 The field of Embodied AI predominantly relies on simulation for training and evaluation, often using either fully synthetic environments that lack photorealism or high-fidelity real-world reconstructions captured with expensive hardware. As a result, sim-to-real transfer remains a major challenge. In this paper, we introduce EmbodiedSplat, a novel approach that personalizes policy training by efficiently capturing the deployment environment and fine-tuning policies within the reconstructed scenes. Our method leverages 3D Gaussian Splatting (GS) and the Habitat-Sim simulator to bridge the gap between realistic scene capture and effective training environments. Using iPhone-captured deployment scenes, we reconstruct meshes via GS, enabling training in settings that closely approximate real-world conditions. We conduct a comprehensive analysis of training strategies, pre-training datasets, and mesh reconstruction techniques, evaluating their impact on sim-to-real predictivity in real-world scenarios. Experimental results demonstrate that agents fine-tuned with EmbodiedSplat outperform both zero-shot baselines pre-trained on large-scale real-world datasets (HM3D) and synthetically generated datasets (HSSD), achieving absolute success rate improvements of 20\% and 40\% on real-world Image Navigation task. Moreover, our approach yields a high sim-vs-real correlation (0.87–0.97) for the reconstructed meshes, underscoring its effectiveness in adapting policies to diverse environments with minimal effort. Project page: \href{https://gchhablani.github.io/embodied-splat/}{https://gchhablani.github.io/embodied-splat}.
\end{abstract}
\section{Introduction}
\label{sec:intro}

Recent advancements in Embodied AI have demonstrated impressive performance in simulated environments \cite{savva2019habitatplatformembodiedai, majumdar2024searchartificialvisualcortex, silwal2024learnlargescalestudypretrained, irshad2021hierarchical, 9956561}. However, translating these capabilities to physical robots remains a significant challenge, primarily due to limitations in simulation fidelity and accessibility \cite{Kadian_2020}. A key bottleneck is the sim-to-real gap, where handcrafted or synthetic simulation environments (e.g. HSSD~\cite{khanna2023habitat}) struggle to capture the complexity and variability of real-world settings, necessitating the use of real-world reconstructions for effective policy training. On the other hand, real-world datasets such as Matterport3D \cite{chang2017matterport3dlearningrgbddata} and HM3D \cite{ramakrishnan2021habitatmatterport3ddatasethm3d}  rely on expensive capture equipment and labor-intensive reconstruction pipelines, making large-scale scene collection and adaptation impractical for many applications. Furthermore, it is difficult to fully capture the variability of potential deployment environments with these datasets.

\begin{figure}[t!]
    \centering
    \includegraphics[width=\linewidth]{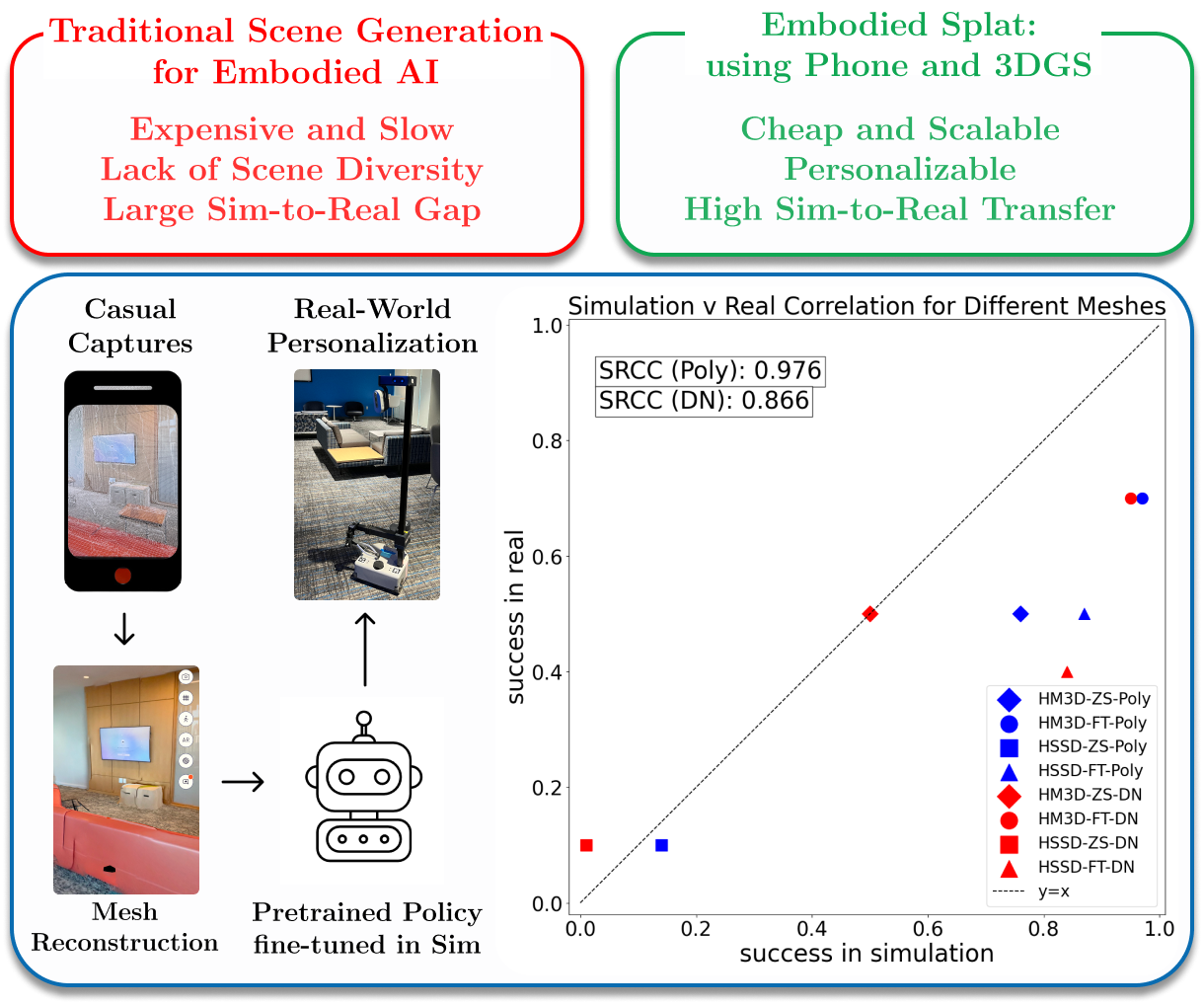}
    \caption{\textbf{Overview of EmbodiedSplat:} Mobile phone captures are used to generate reconstructed meshes via 3D Gaussian Splatting (GS). Agents are trained within these reconstructed environments in simulation before being deployed in the real world, enabling effective sim-to-real transfer. Our analysis demonstrates a strong sim-to-real correlation across both types of generated meshes, highlighting their ability to bridge the gap between simulation and real-world performance.}
    \label{fig:gsnav_teaser}
\end{figure}

Developments in 3D scene representations, particularly 3D Gaussian Splatting (GS)  \cite{kerbl3Dgaussians}, have shown promise towards reducing the effort needed to capture new scenes, enabling high-quality scene reconstruction from casual mobile phone captures. These approaches offer a strong ability to handle complex geometry, perform novel-view view synthesis, and provide high visual fidelity. Building upon this foundation, methods like DN-Splatter \cite{turkulainen2024dnsplatterdepthnormalpriors} have further enhanced mesh reconstruction quality through depth-and-normal regularization. However, their potential for training robot navigation policies and deploying them in the real-world has remained largely unexplored.

In this work, we explore the following question: \textit{Can low-effort cellphone video captures be leveraged to generate meshes that facilitate the training of Embodied AI navigation policies for effective transfer to the target environment?} In other words, we seek to \textit{personalize} models by training them directly on the target distribution, as represented by 3D models built from low-cost and low-effort data collects of the deployment environment. Some recent works indeed explore GS for navigation - GaussNav ~\cite{lei2024gaussnavgaussiansplattingvisual, chen2024splatnavsaferealtimerobot} explores Instance-Image Navigation~\cite{krantz2022instancespecific} in simulation using GS; SplatNav~\cite{chen2024splatnavsaferealtimerobot} explores using GS-based collision meshes for collision avoidance and localization in a single real-world environment using a modular policy for a drone. In contrast, as shown in \cref{tab:comparison_with_other_works}, we explore end-to-end policy learning for image-goal navigation and sim-to-real transfer in indoor environments. To the best of our knowledge, we are the first to explore an approach to solve the personalized real-to-sim-to-real problem using Gaussian Splats for indoor image-goal navigation.

\begin{table}[t!]
\centering
\begin{tabular}{|l|l|c|c|c|}
\hline
\textbf{Work} & \textbf{Goal Type} & \textbf{E2E} & \textbf{Real} & \textbf{S2R} \\ \hline
SplatNav~\cite{chen2024splatnavsaferealtimerobot} & Point/Language & $\times$ & $\checkmark$ & $\times$ \\ \hline
GaussNav~\cite{lei2024gaussnavgaussiansplattingvisual} & Image~\cite{krantz2022instancespecific} & $\times$ & $\times$ & $\times$ \\ \hline
Ours & Image & $\checkmark$ & $\checkmark$ & $\checkmark$ \\ \hline
\end{tabular}
\caption{Comparison against recent works using GS for navigation.``E2E'' indicates whether the policy is trained end-to-end, ``Real'' denotes evaluation on real-world robots, and ``S2R'' indicates demonstrated sim-to-real transfer.}
\label{tab:comparison_with_other_works}
\end{table}

To achieve the above goal, we present a comprehensive framework for leveraging open-source 3D Gaussian Splatting~\cite{kerbl3Dgaussians} (and compare with Polycam~\cite{polycam}). The central premise of our work is that it is possible to quickly capture the scenes in which a robot will be deployed, using readily available consumer-grade hardware, and seamlessly integrate them into simulation environments (i.e. Habitat-Sim~\cite{puig2023habitat}). Policies can then be trained in these simulated environments, leading to improved sim-to-real transfer. Our approach combines the accessibility of smartphone-based scene capture with recent advances in depth-aware 3D scene representation, enabling rapid training and deployment of navigation policies in new, realistic environments.

However, there are several challenges and open questions, ranging from reconstruction quality to finetuning strategies, to enabling successful transfer.
Building on the work of \citet{silwal2024learnlargescalestudypretrained} as a baseline, we test our framework across a range of scenes captured in a university environment, which is out-of-distribution for typical pre-trained policies. We analyze factors affecting transfer performance, including trade-offs between mesh generation pipelines, mesh quality for policy training, and training strategies (e.g., zero-shot vs. fine-tuning). Through rigorous evaluation and real-world robot experiments, we show that our approach yields significant gains in real-world image-goal navigation.

\noindent Our key contributions can be summarized as follows:

\begin{enumerate}
    \item An \textbf{efficient and cost-effective} pipeline for \textbf{bridging the real-to-sim gap in navigation}, enabling the creation of high-quality simulation scenes from low-cost, consumer-grade iPhone captures using depth-aware 3D Gaussian Splats (GS) and Polycam.
    \item \textbf{Comprehensive evaluations} conducted in both simulation and real-world environments, \textbf{assessing zero-shot and fine-tuned policies} on our \cpt scenes. Our results \textbf{demonstrate substantial improvements in sim-to-real transfer}, emphasizing the effectiveness of fine-tuning on high-fidelity reconstructions.
    \item \textbf{In-depth analysis} addressing key research questions, \textbf{exploring the relationship between reconstruction quality, pre-training scenes, downstream navigation performance, and training strategies}. For example, a notable finding is that overfitting policies on a single high-fidelity scene reconstruction in simulation yields reasonable real-world performance. Our findings provide \textbf{valuable insights into how reconstruction fidelity influences policy generalization} and its applicability to real-world scenarios.
    \item An open-source codebase and dataset to facilitate further research in this domain and reproducibility of results.
\end{enumerate}
Through this work, we aim to make high-quality scene collection and agent training more accessible, as well as enable easy development of personalized agents.
\section{Related Work}
\label{sec:related_work}
\subsection{Scene Datasets}
Recent years have seen rapid progress in the development of high-quality 3D scene datasets for embodied AI research. The Matterport3D dataset \cite{chang2017matterport3dlearningrgbddata} provides 10,800 panoramic views from 194,400 RGB-D images across 90 building-scale scenes, complete with surface reconstructions and semantic annotations. Building upon this, HM3D \cite{ramakrishnan2021habitatmatterport3ddatasethm3d} expanded the scale to 1,000 building-scale 3D reconstructions from diverse real-world locations, though some issues with mesh quality were noted. The Gibson dataset \cite{xia2018gibsonenvrealworldperception} introduced another collection of real-world scans, while synthetic datasets like HSSD \cite{khanna2023habitat} (with 211 realistic environments) demonstrated that smaller but higher-quality datasets can sometimes be better than larger ones for agent training. Another dataset that is commonly used is Replica-CAD \cite{szot2021habitat} which provides synthetic variations in layouts of a single scene. In this work, we use HM3D~\cite{ramakrishnan2021habitatmatterport3ddatasethm3d} for training our zero-shot baseline, as it is one of the largest photorealistic indoor-scene datasets captured using Matterport~\cite{matterportCaptureShare} cameras.  Additionally, we use HSSD~\cite{khanna2023habitat} as the synthetic counterpart for another zero-shot baseline. Note, however, that these scenes do have a bias towards apartments, and we therefore test our methods on scenes within a university environment, which is out-of-distribution for these datasets.  The \msr dataset \cite{ren2024mushroommultisensorhybridroom} provides multi-sensor captures of 10 real-world scenes, offering benchmarks for reconstruction and novel view synthesis methods. While this dataset has not been used for embodied tasks in prior works, our dataset collection method is inspired by the \msr dataset's iPhone captures. We evaluate our agents, as well as reconstruction strategies, on these scenes to benchmark our depth and normal encoders.

\subsection{Embodied Navigation in Indoor Environments}

Embodied navigation encompasses various forms of goal-directed navigation tasks~\cite{wijmans2019dd, chaplot2020objectgoalnavigationusing, majumdar2023zsonzeroshotobjectgoalnavigation, alhalah2022zeroexperiencerequiredplug, yadav2023ovrlv2simplestateofartbaseline, yadav2022offlinevisualrepresentationlearning,krantz2022instancespecific, bono2023endtoendinstanceimagegoalnavigation}. This study primarily investigates ImageNav~\cite{alhalah2022zeroexperiencerequiredplug, yadav2023ovrlv2simplestateofartbaseline, yadav2022offlinevisualrepresentationlearning} in indoor environments. Recent advancements in image-goal navigation have explored a variety of approaches to enhance performance and generalization. Notably, pre-trained visual representations~\cite{yadav2022offlinevisualrepresentationlearning, yadav2023ovrlv2simplestateofartbaseline, majumdar2024searchartificialvisualcortex, silwal2024learnlargescalestudypretrained} have shown substantial promise in improving performance on the ImageNav task. In particular, \citet{silwal2024learnlargescalestudypretrained} demonstrate remarkable zero-shot success rates (90\%) on real-world ImageNav, leveraging VC-1~\cite{majumdar2024searchartificialvisualcortex}, fine-tuned end-to-end with data augmentation. Inspired by this, we use the setup in ~\citet{silwal2024learnlargescalestudypretrained} for our policy training and evaluation.

Sim-to-real transfer remains a core challenge in Embodied AI beyond leveraging pre-trained visual encoders. Kadian et al.~\cite{Kadian_2020} proposed the Sim-vs-Real Correlation Coefficient (SRCC) to quantify how well PointNav performance in simulation predicts real-world results. We build on this idea to evaluate the sim-to-real predictivity of our \cpt meshes.

A closely related work is Phone2Proc~\cite{deitke2022phone2procbringingrobustrobots}, which demonstrates enhanced sim-to-real ObjectNav performance by generating layouts from iPhone RoomPlan API captures. While similar to our approach in utilizing iPhone captures, our methodology differs in that we capture the entire room without focusing on layouts, nor do we perform any post-processing of scenes or generate multiple variations of the same scene for agent training. Instead, we generate meshes from our captures, fine-tune pre-trained policies for ImageNav task, and deploy them subsequently.

\subsection{3D Scene and Mesh Reconstruction}

\looseness=-1 In this work, we adopt DN-Splatter~\cite{turkulainen2024dnsplatterdepthnormalpriors} for its simplicity and superior performance (and compare to meshes from  Polycam~\cite{polycam}, which is not open-source). DN-Splatter uses depth-and-normal regularization to improve mesh quality. While mesh reconstruction has also been explored with NeRFs~\cite{mildenhall2020nerfrepresentingscenesneural}, we choose GS for 3D reconstruction due to their fast training and rendering speeds, improving the overall efficiency of the pipeline. For an expanded list of related works on this topic, please see \cref{subsec: appendix_related_work_3d_scene}.

\subsection{3D Representations for Embodied AI/Robotics}
There have been works which explore sim-to-real manipulation~\cite{byravan2022nerf2realsim2realtransfervisionguided, wu2024rlgsbridge3dgaussiansplatting, li2024objectaware}. In this work, we attempt to solve the problem of sim-to-real navigation, laying the foundation for a diverse range of embodied tasks, such as object-goal navigation, rearrangement, and mobile manipulation, which require room-, apartment-, and building-level scene representations. Capturing high-quality 3D Gaussian Splats (GS) is relatively straightforward for tabletop manipulation tasks, whereas large-scale scene reconstruction poses greater challenges — a key aspect of this work.
Recent works also explore the use of GS in solving the embodied navigation problem. GaussNav~\cite{lei2024gaussnavgaussiansplattingvisual} presents a semantic Gaussian-based map reconstruction for the HM3D-Instance ImageNav~\cite{krantz2022instancespecific} task. Splat-Nav~\cite{chen2024splatnavsaferealtimerobot} introduces a two-stage pipeline for planning and pose estimation through Gaussian Splat Maps. \cref{tab:comparison_with_other_works} highlights the differences between our approach and recent works utilizing GS for navigation. By capturing the entire scene using mobile phones, we enable personalization to specific scenes, as well as the potential to scale scene collection in the future. Some works~\cite{marza2023autonerftrainingimplicitscene} explore automated 3D scene capture using robot cameras in simulation, but these methods require complex, embodiment-dependent strategies that are time-consuming and difficult to scale. In contrast, our approach leverages human captures in the real-world, offering a fast and scalable solution without the need for intricate planning.
For more related work on this topic, please see \cref{subsec: appendix_related_work_3d_scene_in_robotics}.
\section{Methodology}
\label{sec:methodology}

\begin{figure*}[h!]
    \centering
    \includegraphics[width=1\linewidth]{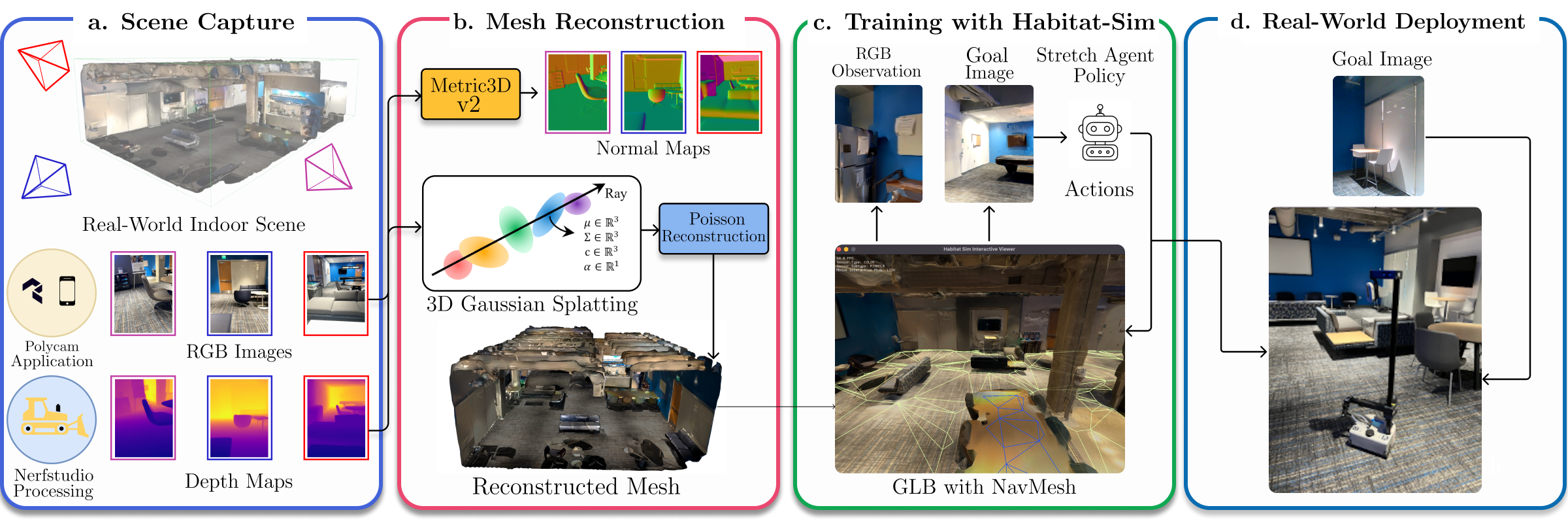}
    \caption{\textbf{The EmbodiedSplat Pipeline}: Pipeline for integration of real-world captures with Habitat-Sim~\cite{puig2023habitat} and subsequent deployment. The first stage (a.) involves capturing the scene using Polycam~\cite{polycam} and Nerfstudio~\cite{nerfstudio} which produces RGB frames, associated iPhone GT depth maps, and poses. In the second stage (b.), we use DN-Splatter~\cite{turkulainen2024dnsplatterdepthnormalpriors} to train GS using depth and normal regularization, with normals from Metric3D-V2~\cite{hu2024metric3d} monocular encoder. A mesh (\texttt{.ply}) is created using Poisson reconstruction from the GS. In the third stage (c.), the mesh is processed and loaded into Habitat-Sim~\cite{puig2023habitat} for training the agent in simulation. In the last stage (d.), the policy is deployed in the real-world in the same scene for image-goal navigation.
    }
    \label{fig:gsnav_pipeline}
\end{figure*}

\looseness=-1 The overall pipeline for bridging and integrating a real-world scene with Habitat-Sim~\cite{puig2023habitat} is shown in \cref{fig:gsnav_pipeline}. 
In the following subsections, we discuss each stage of the pipeline in detail. \cref{subsec:datasets} discusses the datasets used and the details of scene captures. \cref{subsec:mesh_reconstruction} discusses the second stage, where these captures are converted to meshes. \cref{subsec:imagenav_episode} discusses how ImageNav episodes are created, which is a crucial part of integrating the scene with Habitat Simulator~\cite{puig2023habitat}. \cref{sec:appendix_training_details} discusses agent training and evaluation details. For details on real-world deployment setup, refer to \cref{sec:appendix_real_world_methodology}.

\subsection{Datasets and Scene Captures}
\label{subsec:datasets}
\noindent \textbf{Scene Datasets:}
We use two datasets for pre-training ImageNav policies, which serve as zero-shot baselines and pre-trained policies for our scenes, both in simulation and real-world evaluations. Specifically, we use the HM3D~\cite{ramakrishnan2021habitatmatterport3ddatasethm3d} dataset which consists of apartment-scale scenes split into 800 training and 100 validation scenes, and the HSSD~\cite{khanna2023habitat} dataset split into 134 training and 33 validation scenes. 

\noindent \textbf{\cpt Scenes:}
To evaluate the feasibility of the pipeline and conduct real-world experiments, we capture scenes from a university environment (classroom, community lounges, conference rooms, etc). For custom data collection with an iPhone, we follow the procedure used for collecting the \msr dataset~\cite{ren2024mushroommultisensorhybridroom}. Specifically, we use an iPhone 13 Pro Max to record the iPhone RGB-D data using the Polycam application~\cite{polycam}. Polycam provides an assistive interface to ensure that all the details of the scene have been sufficiently covered during the capture. We use the default settings with Polycam and export the raw data exposed by the application. Subsequently, we use Nerfstudio~\cite{nerfstudio} to process the RGB-D data and sample 1000 aligned RGB-depth frames with low blur scores and corresponding poses. Additionally, Polycam also provides a mesh with its exported data. We use this mesh for comparison purposes, referred to as \poly meshes throughout the paper.
Unlike \citet{ren2024mushroommultisensorhybridroom}, we use a manually-held iPhone for capture instead of a gimbal, enabling us to evaluate whether the process remains effective and easily replicable without relying on expensive capture equipment. Each capture requires 20-30 minutes of recording with Polycam. We repeat this process for different indoor scenes, which we refer to as \lng, \csr, \pid, and \cab. The scale of these
of these scenes is similar to those in the \msr dataset (1-3 rooms) ~\cite{ren2024mushroommultisensorhybridroom}.

\noindent \textbf{\msr Dataset:}
We use \msr ~\cite{ren2024mushroommultisensorhybridroom} dataset for benchmarking. The dataset consists of long and short sequences of 10 indoor scenes, captured with iPhone, Kinect, and Faro Scanner. In our work, we only use the iPhone long-sequences for training and evaluation purposes. We benchmark different normal encoders on all 10 scenes (see \cref{sec:appendix_depth_normals}) for DN-Splatter~\cite{turkulainen2024dnsplatterdepthnormalpriors}. For agent training and evaluation, we primarily use three scenes - \texttt{honka}, \texttt{sauna}, \texttt{activity} - which have varied scale and complexity.

\subsection{3D Scene and Mesh Reconstruction}
\label{subsec:mesh_reconstruction}

\begin{figure*}[ht!]
    \centering
    \begin{subfigure}{0.24\textwidth}
        \centering
        \includegraphics[width=\textwidth]{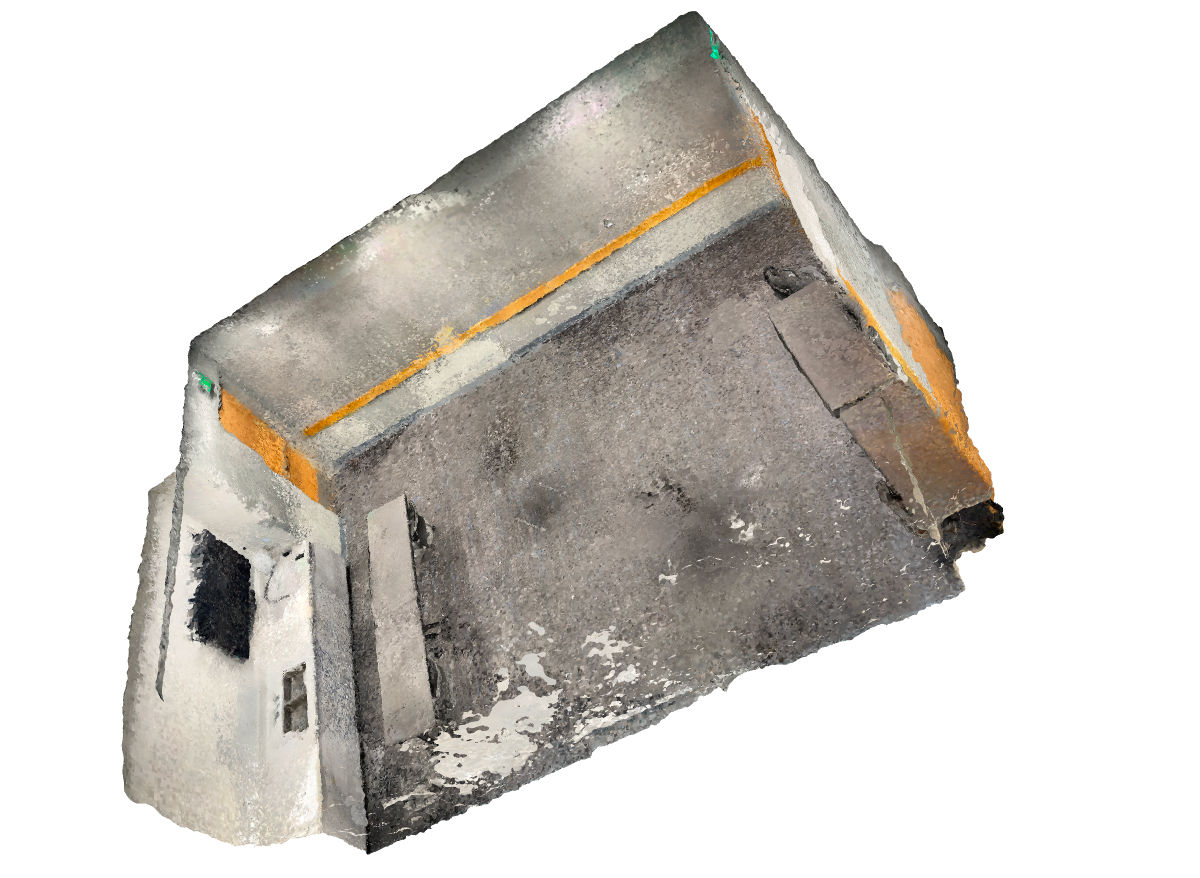}
        \caption{\cab}
        \label{fig:castleberry}
    \end{subfigure}
    \hfill
    \begin{subfigure}{0.24\textwidth}
        \centering
        \includegraphics[width=\textwidth]{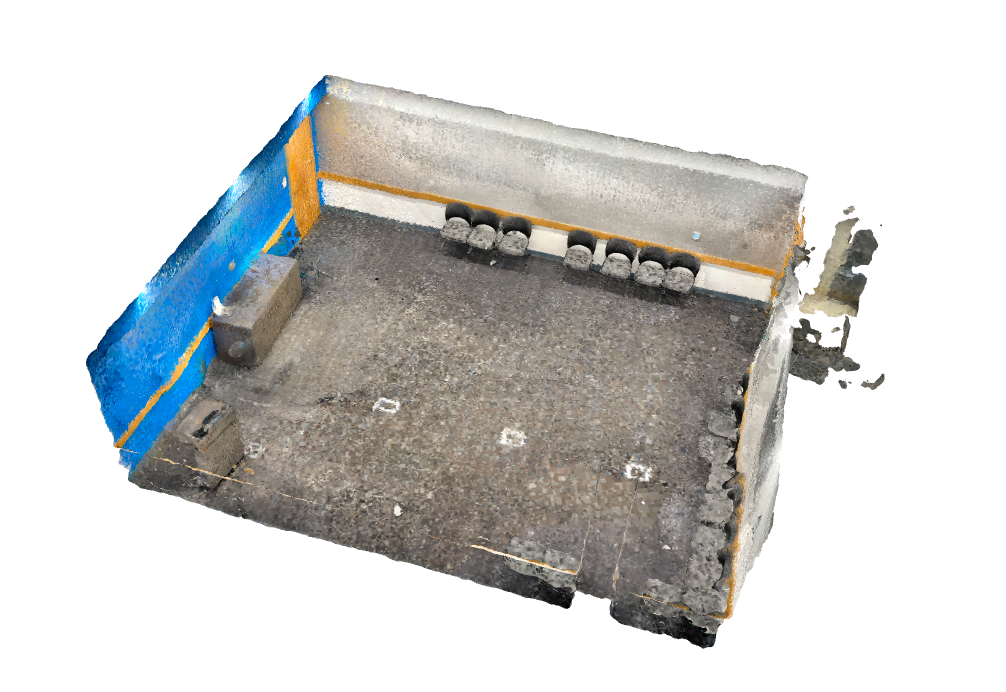}
        \caption{\pid}
        \label{fig:piedmont}
    \end{subfigure}
    \hfill
    \begin{subfigure}{0.24\textwidth}
        \centering
        \includegraphics[width=\textwidth]{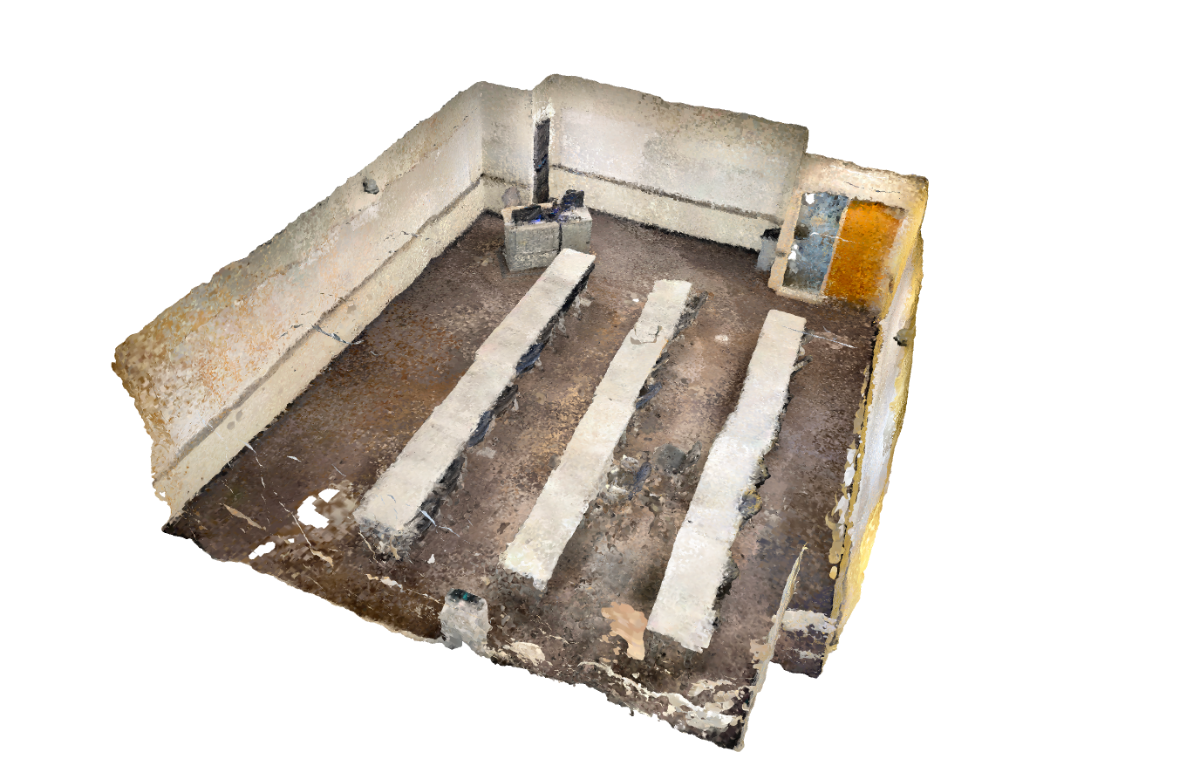}
        \caption{\csr}
        \label{fig:classroom}
    \end{subfigure}
    \hfill
    \begin{subfigure}{0.24\textwidth}
        \centering
        \includegraphics[width=\textwidth]{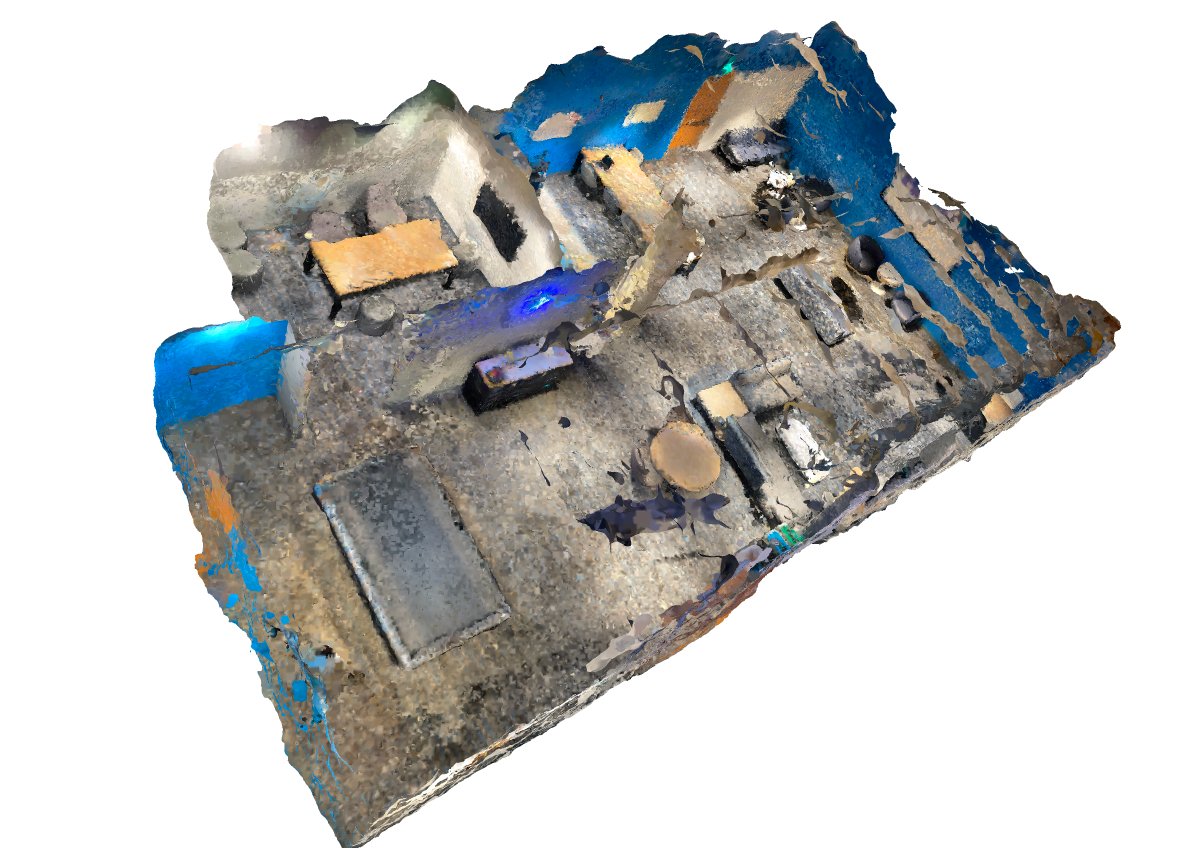}
        \caption{\lng}
        \label{fig:grad_lounge}
    \end{subfigure}
    \caption{Reconstructed meshes of the \cpt scenes after DN-Splatter~\cite{turkulainen2024dnsplatterdepthnormalpriors} training and Poisson reconstruction.}
    \label{fig:mesh_examples}
\end{figure*}

For mesh reconstruction, we use DN-Splatter~\cite{turkulainen2024dnsplatterdepthnormalpriors} as our method of choice for its superior performance on mesh reconstruction in comparison to others, and simplicity of its integration with Habitat-Sim~\cite{puig2023habitat}. DN-Splatter leverages depth-normal regularization, and smoothness losses to maintain geometric consistency during Gaussian Splat training. For mesh reconstruction, the rendered depth and normal maps are back-projected from training views to create a point-set for meshing.

We use the default hyperparameters from DN-Splatter~\cite{turkulainen2024dnsplatterdepthnormalpriors} and train the GS for 30,000 iterations. We use sensor depth $\lambda_{d} = 0.2$, and enable the depth smoothness and normal losses. Metric3D-V2~\cite{hu2024metric3d} is used as the normal encoder instead of Omnidata~\cite{eftekhar2021omnidata, kar20223d}, as empirical results demonstrate it yields higher-quality meshes. For further discussion on depth and normal encoder selection, see \cref{sec:appendix_depth_normals}.

After training GS and converting into \texttt{ply} meshes, we convert the meshes to \texttt{glb} in Blender (see \cref{sec:appendix_mesh_processing}) before loading them in Habitat-Sim~\cite{puig2023habitat}. The final meshes for our \cpt scenes are shown in \cref{fig:mesh_examples} and referred to as \dn mesh hereafter to contrast against the \poly mesh counterpart. It takes approximately 20-30 minutes per capture, and 1-2 hours of training with DN-Splatter~\cite{turkulainen2024dnsplatterdepthnormalpriors} to generate these meshes, which is significantly lesser compared to the cost and several hours of capture and processing with Matterport~\cite{matterportCaptureShare} cameras. 

\subsection{ImageNav Episode Generation}
\label{subsec:imagenav_episode}
\noindent \textbf{Pre-training Scene Datasets: }
For the HM3D~\cite{ramakrishnan2021habitatmatterport3ddatasethm3d} and HSSD~\cite{khanna2023habitat} datasets, we use the predefined train-validation split, generating 10,000 episodes for each training scene and 25 episodes for each validation scene (see \cref{subsec:appendix_generate_imagenav_episodes}).

\noindent \textbf{\cpt and \msr Scenes:}
For scenes with \dn meshes and \poly meshes, we select only the largest navmesh island within each scene. We ensure that the largest island covers most, if not all, of the navigable areas in the scene.  This step is crucial due to the relatively smaller sizes of these scenes compared to HM3D scenes. Hence, we generate only 1000 training episodes and 100 evaluation episodes per scene. By selecting the largest navmesh island, we minimize the risk of incorrectly treating entities other than the floor as navigable.

\noindent \textbf{Evaluation Metric:} We consider Success Rate (SR), the fraction of successful episodes out of all the episodes, as the primary metric for performance evaluation of our policies. An episode is considered successful if the agent stops within 1m of the goal location before the maximum number of steps (1000 for simulation, 100 for real) are over.
\section{Experiments}
\label{sec:experiments}
This section aims to answer the following research questions:
\begin{enumerate}
    \item How does a pre-trained policy perform on \cpt and \msr scenes? (\cref{subsec:zero_shot})
    \item Does fine-tuning on the \cpt and \msr scenes improve performance? (\cref{subsec:fine_tune}).
    \item Does the performance in simulation transfer to the real-world? (\cref{subsec:real_world_transfer}).
\end{enumerate}
Further experiments and analyses are discussed in \cref{sec:ablations}.

\subsection{How does a pre-trained policy perform on \cpt and \msr scenes?}
\label{subsec:zero_shot}

In this experiment, we evaluate the pre-trained policies that achieved the highest validation success rates on the HM3D~\cite{ramakrishnan2021habitatmatterport3ddatasethm3d} (83.08\% val SR) and HSSD~\cite{khanna2023habitat} (63.15\% val SR), trained for 600M and 1200M steps, respectively. Note that the performance on HM3D is consistent with the simulation results reported in~\citet{silwal2024learnlargescalestudypretrained} and represents the current state-of-the-art for this task. For agent embodiment, policy training and evaluation details, please refer to \cref{sec:appendix_training_details}. We assess the zero-shot generalization of these policies within simulation, evaluating their performance across different meshes for individual scenes.

\begin{figure}[h!]
\centering
\includegraphics[width=1\linewidth]{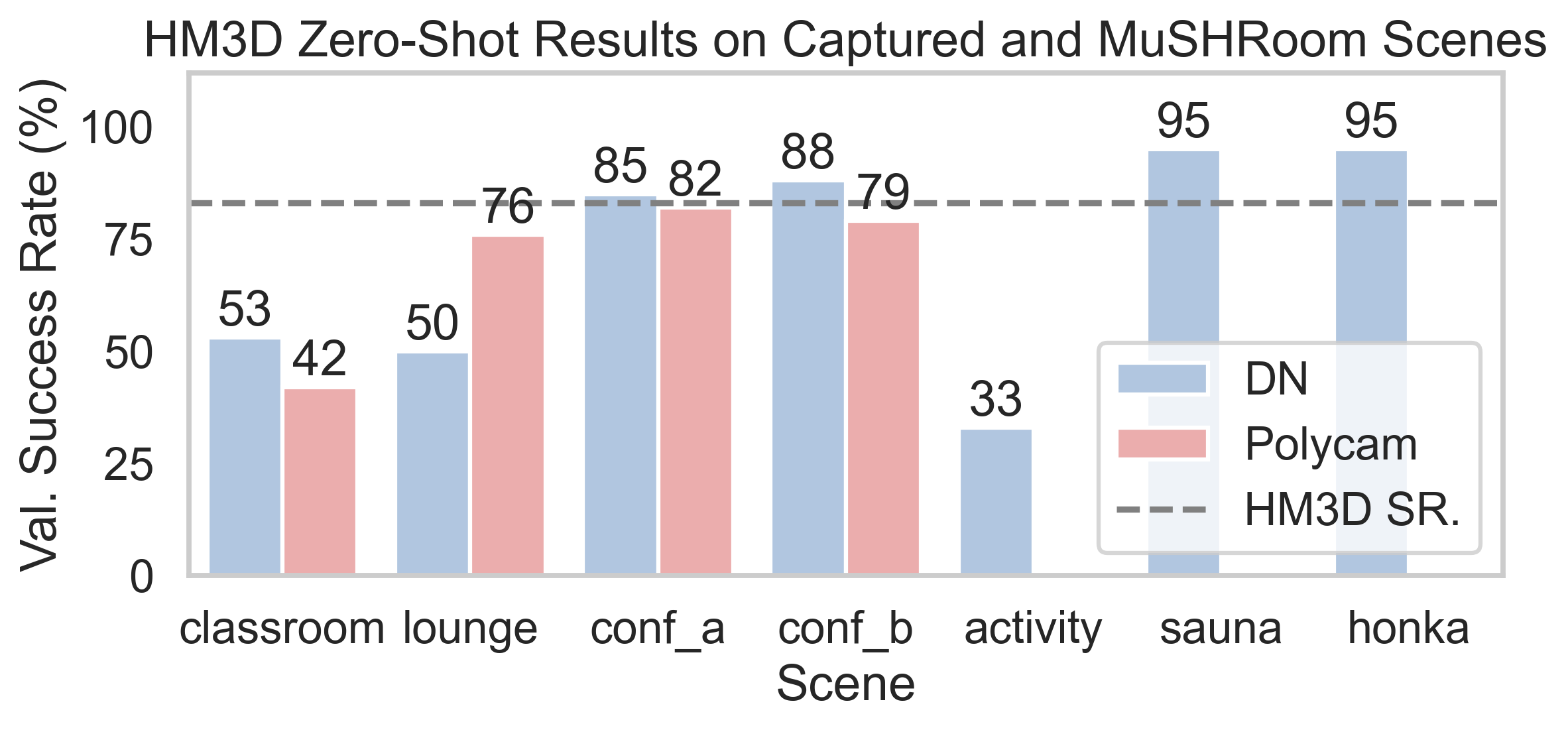}
\caption{\textbf{Zero-shot val. SR for HM3D pre-trained policy}.}
\label{fig:zero-shot-hm3d}
\end{figure}

\begin{figure}[h!]
\centering
\includegraphics[width=1\linewidth]{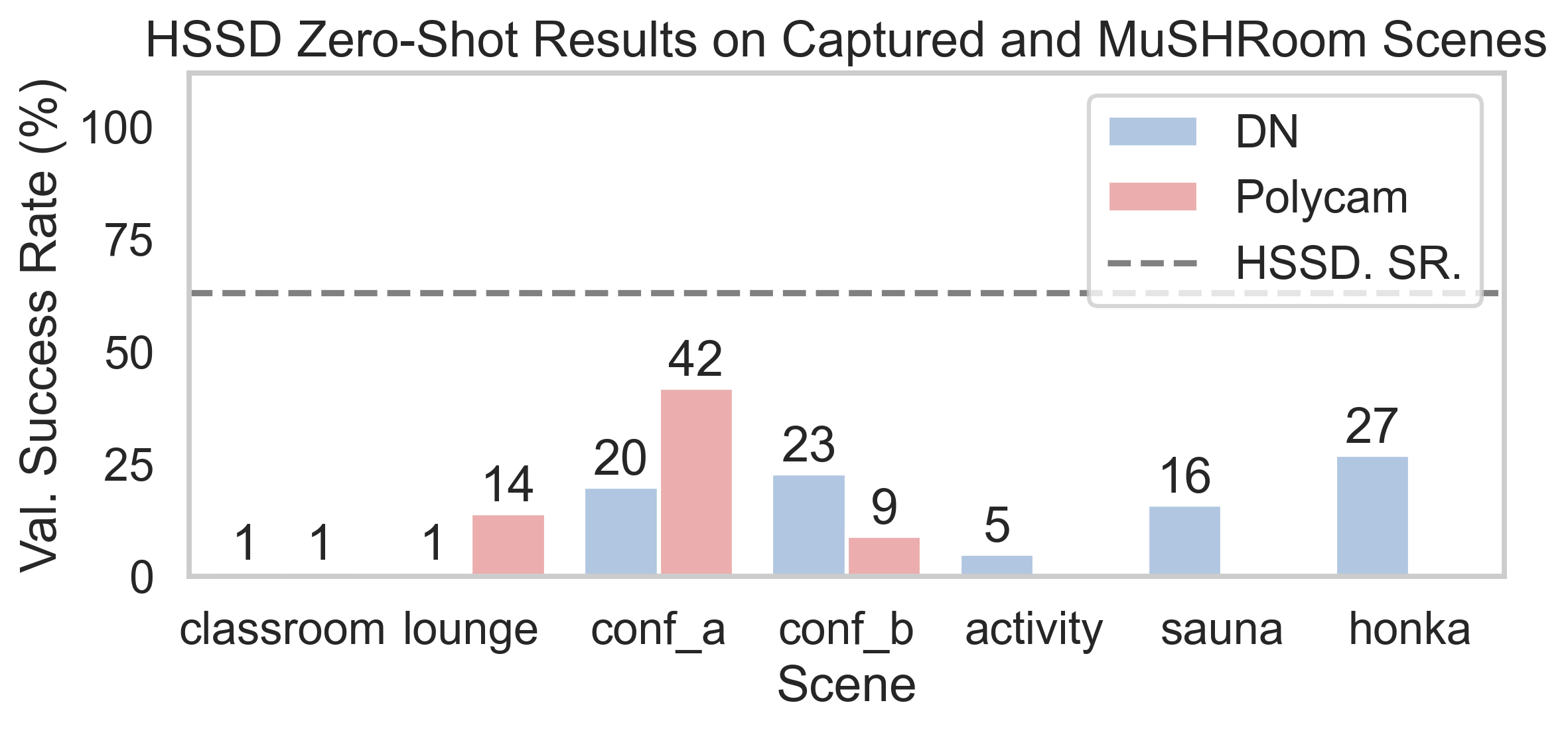}
\caption{\textbf{Zero-shot val. SR for HSSD pre-trained policy}.}
\label{fig:zero-shot-hssd}
\end{figure}

\cref{fig:zero-shot-hm3d} presents the results of the zero-shot evaluation across individual scenes for the HM3D pre-trained policy. Performance varies significantly based on scene size, complexity, and mesh type. For smaller scenes such as \pid and \cab, the policy demonstrates relatively high success rates in both \dn mesh and \poly settings, achieving 85\% (\dn) and 82\% (\poly) for \pid and 88\% (\dn mesh) and 79\% (\poly) for \cab.

In contrast, performance declines in larger environments such as \csr and \lng. In \csr, success rates drop to 53\% with \dn mesh and 42\% with \poly, while in \lng, the policy achieves 50\% (\dn mesh) and 76\% (\poly). These results highlight the challenge of transferring the policy to more complex, large-scale environments that differ from the training data. Since HM3D consists primarily of apartment-style scenes, the policy has not been exposed to classrooms or community lounges during training. Both \dn and \poly meshes yield similar success rates on average($\sim60\%$), suggesting no clear advantage of one mesh type over the other in aligning with HM3D.
Additionally, we evaluate the policies on three scenes from the \msr ~\cite{ren2024mushroommultisensorhybridroom} dataset. Due to the unavailability of \poly meshes for \msr, we conduct evaluations solely on \dn meshes. The policy achieves a 95\% success rate in the \texttt{sauna} and \texttt{honka} scenes, suggesting promising potential for performance improvements through additional scene refinements (such as using a gimbal during capture). However, in the larger \texttt{activity} scene, performance declines, indicating challenges in generalizing to larger environments.

\looseness=-1 \cref{fig:zero-shot-hssd} reveals similar trends for the HSSD pre-trained policy, though with significantly lower success rates overall. This degradation in performance can be attributed to the synthetic nature of HSSD scenes, which lack the realism and scale necessary for effective generalization. The policy performs particularly poorly on \csr, achieving only 1\% success on both \dn mesh and \poly, while performance in \lng is slightly better, reaching 14\% success with \poly. The policy exhibits slightly improved performance on other scenes, but it is hard to conclude which mesh type best aligns with HSSD.

\subsection{Does fine-tuning on the \cpt and \msr scenes improve performance?}
\label{subsec:fine_tune}

\begin{figure}[t]
\centering
\includegraphics[width=1\linewidth]{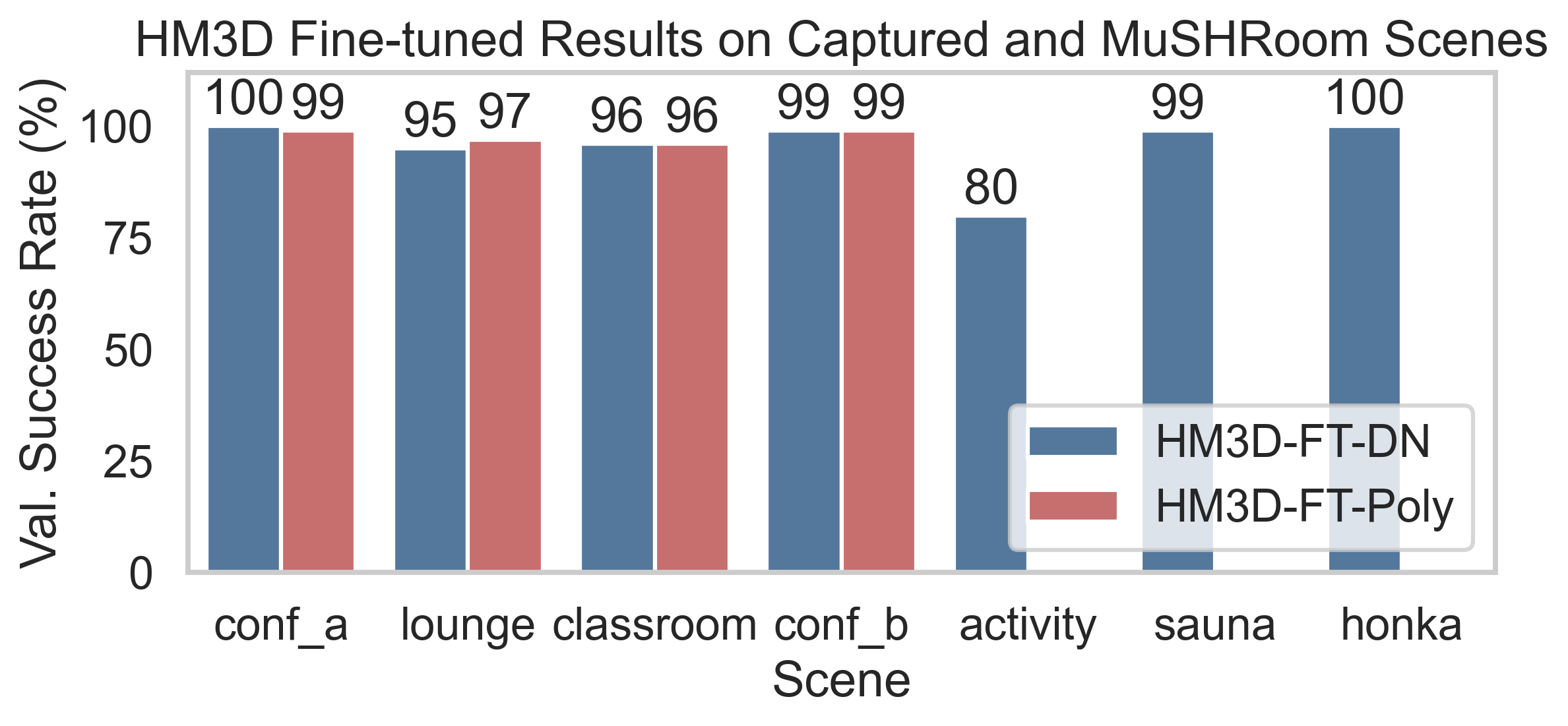}
\caption{\textbf{Fine-tuned validation SR from HM3D pre-trained policy}: The policy is fine-tuned and evaluated on the same mesh.}
\label{fig:fine-tune-hm3d}
\end{figure}

\begin{figure}[t]
\centering
\includegraphics[width=1\linewidth]{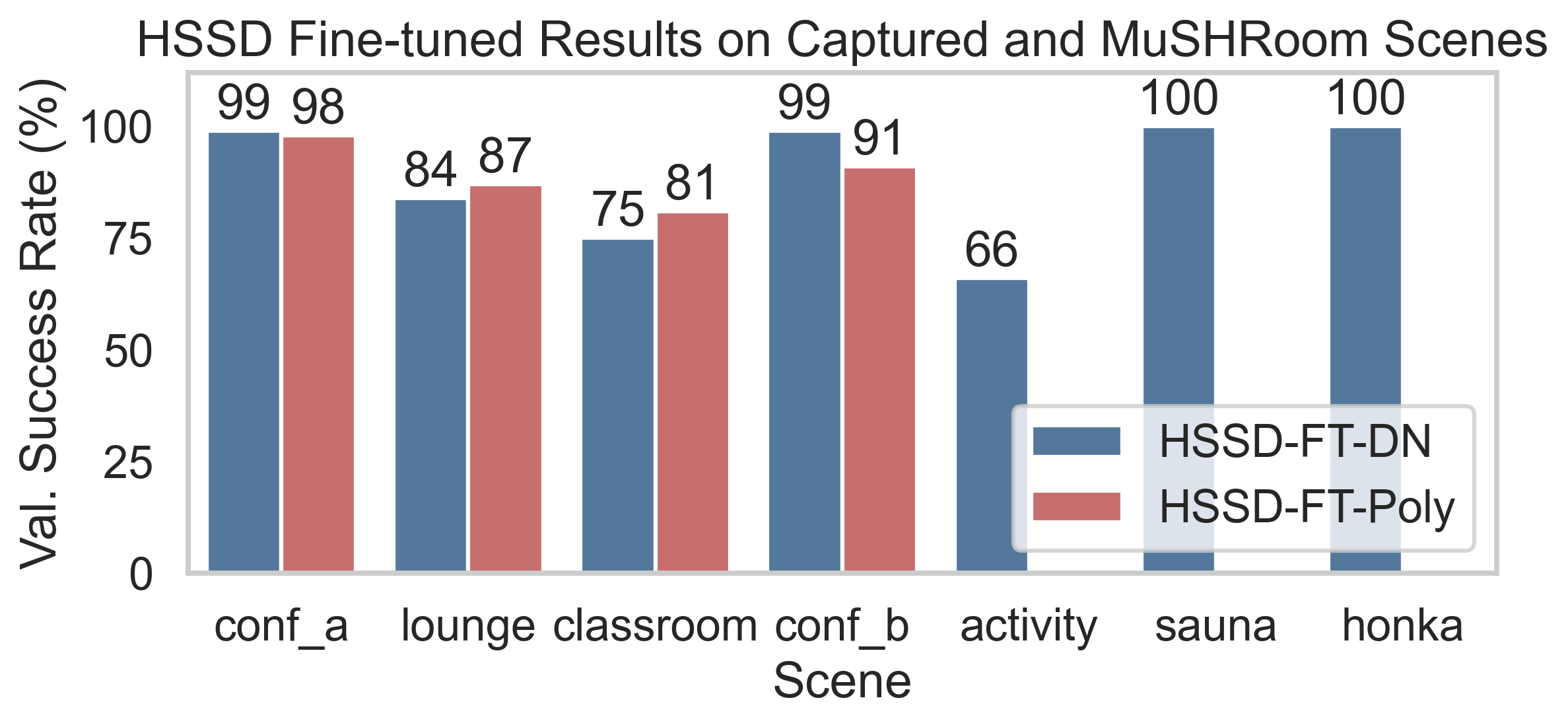}
\caption{\textbf{Fine-tuned validation SR from HSSD pre-trained policy}: The policy is fine-tuned and evaluated on the same mesh.}
\label{fig:fine-tune-hssd}
\end{figure}

\looseness=-1 In this section, we investigate whether fine-tuning the pre-trained policies for only 20M additional steps improves performance in simulation. The learning rate is set to $2.5e{-6}$ for the LSTM policy and $6e{-7}$ for the visual encoder, following a fine-tuning strategy similar to that of~\citet{deitke2022phone2procbringingrobustrobots}. We fine-tune the pre-trained policy on the training episodes for a single scene and evaluate on the corresponding validation episodes.

The results are presented in \cref{fig:fine-tune-hm3d} and \cref{fig:fine-tune-hssd}.  We observe that fine-tuning significantly improves performance across all tested scenes. For the pre-trained HM3D policy, fine-tuning for 20M steps results in success rates approaching 90\%+ across different meshes. Similarly, for the HSSD pre-trained policy, fine-tuning leads to substantial improvements, with most policies achieving success rates of 80\%+ in respective scenes. In particular, performance gains are particularly pronounced in larger, more complex environments such as \csr and \lng, which differ significantly from apartment-style scenes in HM3D and HSSD. These results indicate that our pipeline can effectively be used to collect diverse scene data at scale. Furthermore, they suggest that agents can be quickly fine-tuned (only 20M steps vs 100M+ steps for training from scratch) on these generated meshes to improve performance on specific scene types, enhancing personalization beyond the original training distribution. We present results on additional scenes in \cref{sec:additional_results}.

\subsection{Does the performance in simulation transfer to the real-world?}
\label{subsec:real_world_transfer}

We evaluate both zero-shot and fine-tuned policies in the real-world \lng scene on a Stretch robot. During the evaluation, each episode is capped at 100 steps, with 10 distinct start-and-goal locations sampled within the scene (see \cref{sec:appendix_real_world_methodology} for deployment details). To assess performance, we record the number of steps taken and final distance to the goal at the end of each episode, determining whether the agent successfully completes the task.

\begin{figure}[h]
\centering
\includegraphics[width=1\linewidth]{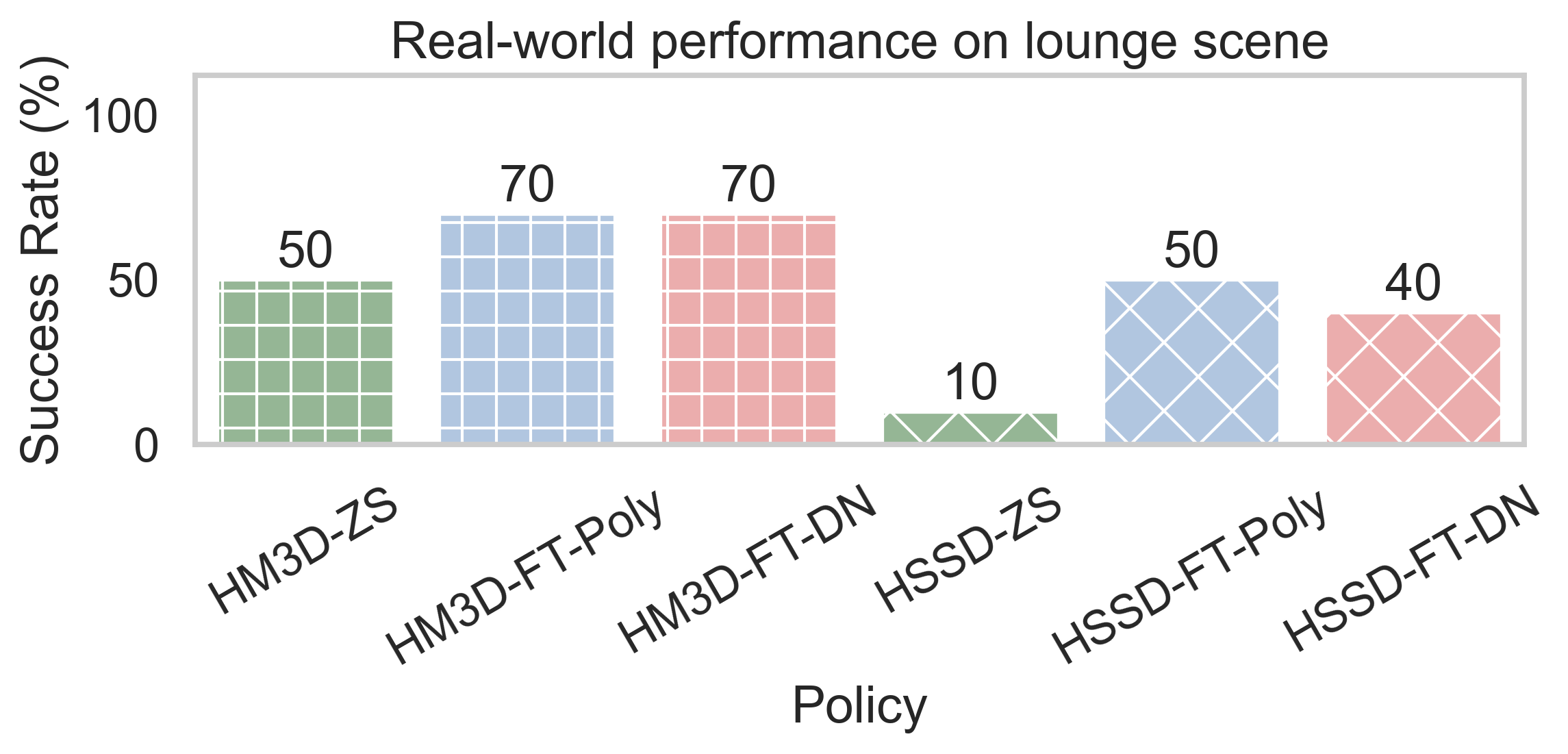}
\caption{\textbf{Real world results of zero-shot and fine-tuned models on \lng scene}. HM3D-ZS (real dataset) leads to a better success rate than HSSD-ZS (synthetic dataset). We observe increased real-world success with fine-tuning on \dn and \poly meshes for both HM3D and HSSD policies.}
\label{fig:real_world_results}
\end{figure}

\cref{fig:real_world_results} presents results across 10 evaluation episodes. The zero-shot HM3D policy achieves a 50\% success rate, demonstrating our hypothesized lack of generalization. This is in contrast with the results reported in~\citet{silwal2024learnlargescalestudypretrained} showing 90\% zero-shot success rate in the real-world. We attribute this discrepancy to the structural and semantic differences between the \lng and the apartment-style scenes typically encountered in HM3D. Fine-tuning on the \poly and \dn mesh reconstructions of this scene improves performance, with success rates increasing to 70\%. For HSSD, zero-shot performance is significantly lower at 10\%, while fine-tuned policies improve success rates to 50\% with \poly and 40\% with \dn mesh.

\cref{fig:gsnav_teaser} illustrates the Sim-to-Real Correlation Coefficient (SRCC)~\cite{Kadian_2020} between simulation and real-world performance. The observation suggests that improvements in evaluation performance on \dn and \poly meshes in simulation translate to improved real-world performance. This demonstrates that our approach can efficiently adapt policies to novel real-world environments. We discuss real-world statistics (number of steps, distance) in \cref{sec:appendix_real_world_metrics}.
\section{Ablations and Analysis}
\label{sec:ablations}
In this section, we systematically investigate critical factors influencing policy performance. First, we assess whether pre-training on large-scale datasets such as HM3D or HSSD is necessary for effective real-world transfer (\cref{subsec:overfit_results}). Next, we analyze the impact of validation PSNR and scene geometry on the zero-shot performance (\cref{subsec:val_psnr_sr_dist}). Finally, we examine whether continuous training on large-scale datasets improves zero-shot performance on our \cpt scenes, shedding light on the relationship between dataset characteristics and generalizability (\cref{subsec:line_chart_success_rates}).
 
\subsection{Is it necessary to pre-train over large-scale datasets?}
\label{subsec:overfit_results}

\begin{figure}[h]
\centering
\includegraphics[width=1\linewidth]{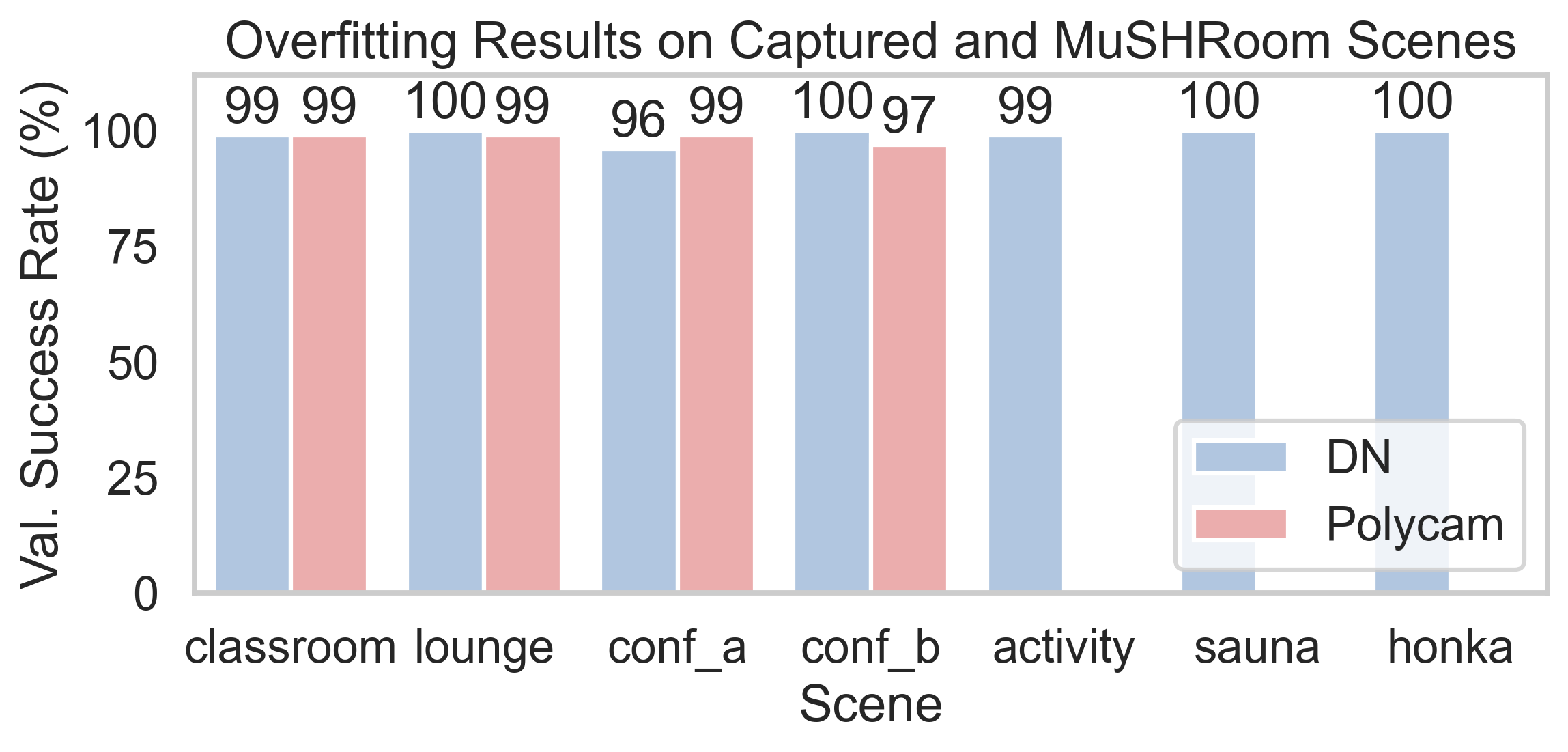}
\caption{\textbf{Validation SR for overfitted policies on respective scenes}: Polices are not pre-trained on HM3D/HSSD.}
\label{fig:overfit_combined}
\end{figure}

\looseness=-1 In this experiment, we investigate the necessity of pre-training on large-scale datasets by ``overfitting'' policies directly on \poly and \dn meshes  with a policy trained from scratch (not pre-trained on large-scale datasets such as HM3D or HSSD) for $\sim$100M steps. Training and validation data are derived from ImageNav episode generation, as described in \cref{subsec:imagenav_episode}. To ensure diversity, start-goal locations for training and evaluation episodes are randomly sampled. The best validation success rates for all policies on their respective scenes are presented in \cref{fig:overfit_combined}. As expected, the overfitted policies achieve near-perfect performance in simulation, maintaining high success rates. For a detailed discussion on why overfitting results seem better than fine-tuning in this scenario, please see \cref{sec:appendix_overfit_vs_finetune}.

We further evaluate these overfitted policies in the real-world \texttt{lounge} scene using \poly and \dn meshes. Surprisingly, the overfitted policy trained on the \poly mesh achieves a 50\% success rate in real-world evaluations. In contrast, the policy trained on the \dn mesh achieves only 10\% success. This result suggests that our mesh-based training approach can yield non-zero real-world performance, even without large-scale pre-training. We attribute the performance gap between \poly and \dn-trained policies to differences in visual fidelity—\poly meshes preserve more visual detail by directly utilizing original images to reconstruct the scene, whereas \dn meshes are based on GS which use the learned colors for the 3D Gaussians.

\subsection{How is the scale and validation PSNR of the scene related to the final performance?}
\label{subsec:val_psnr_sr_dist}

\begin{figure}[h]
\centering
\includegraphics[width=1\linewidth]{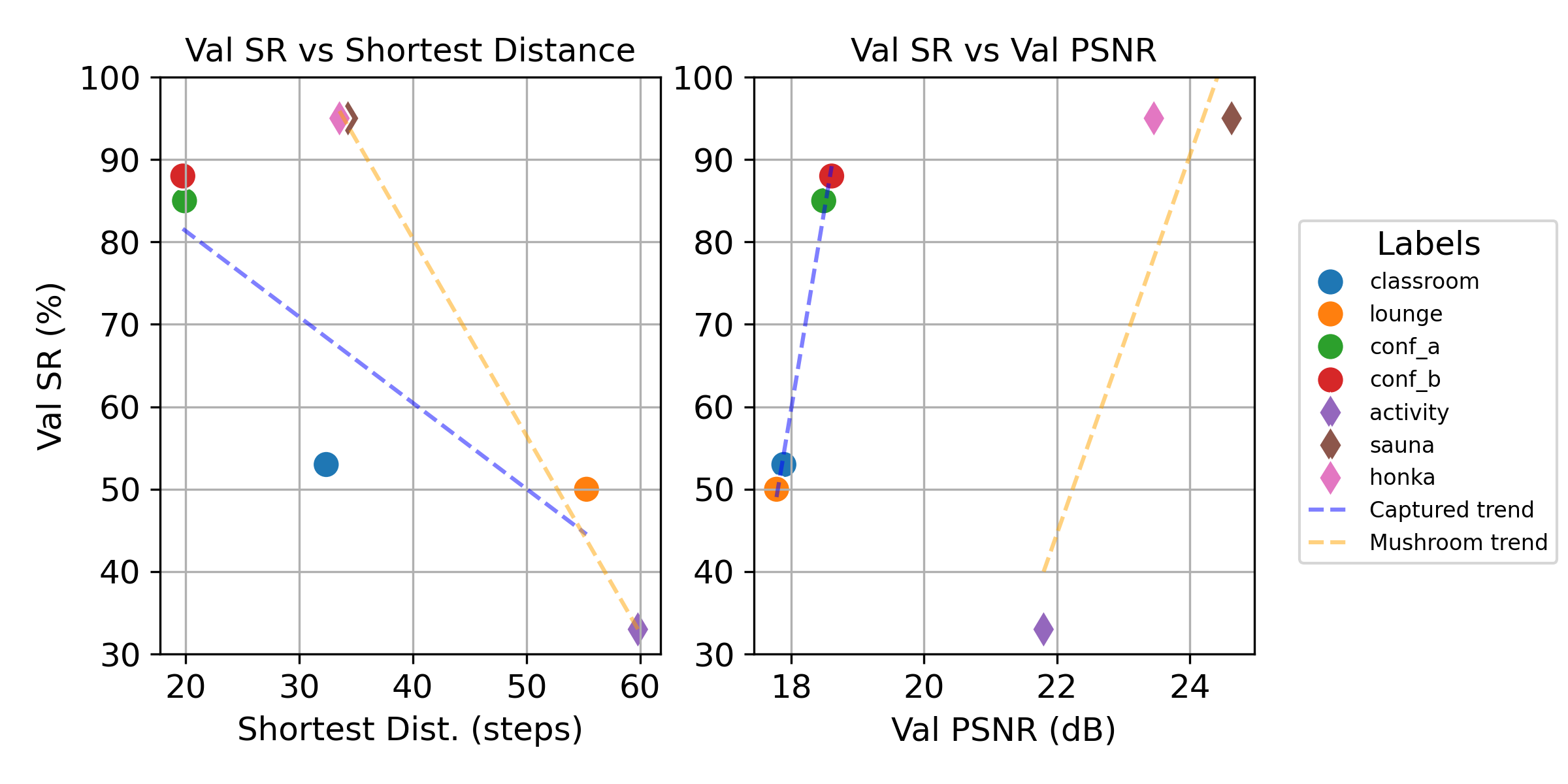}
\caption{\textbf{HM3D zero-shot validation success rates on \dn meshes vs. validation GS PSNRs vs. average shortest distances of validation episodes}. The zero-shot SR is inversely correlated with the scale of the scene, while directly correlated with val. PSNR on the trained GS.}
\label{fig:val_sr_vs_psnr_vs_shortest_dist}
\end{figure}

\looseness=-1 \cref{fig:val_sr_vs_psnr_vs_shortest_dist} illustrates the correlation between HM3D zero-shot validation success rates on \dn meshes and two key factors: the Peak Signal-to-Noise Ratio (PSNR) of the corresponding 3D Gaussian Splats (GS) and the scale of the scene, measured as the average shortest distance between start-goal locations in validation episodes. We observe a negative correlation between success rate (SR) and average shortest distance—indicating that as the scale of the scene increases, the zero-shot success rate declines. Conversely, a positive correlation is seen between SR and PSNR, where higher validation PSNR values correspond to improved success rates. Notably, different trend lines emerge for \msr captures and our own \cpt scenes. \msr captures generally exhibit higher validation PSNRs, likely due to the use of a stabilized gimbal during data collection. Future work will further investigate the impact of capture stability on policy performance.

\subsection{Does continuous training on large-scale datasets improve zero-shot performance?}
\label{subsec:line_chart_success_rates}

\begin{figure}[h]
\centering
\includegraphics[width=1\linewidth]{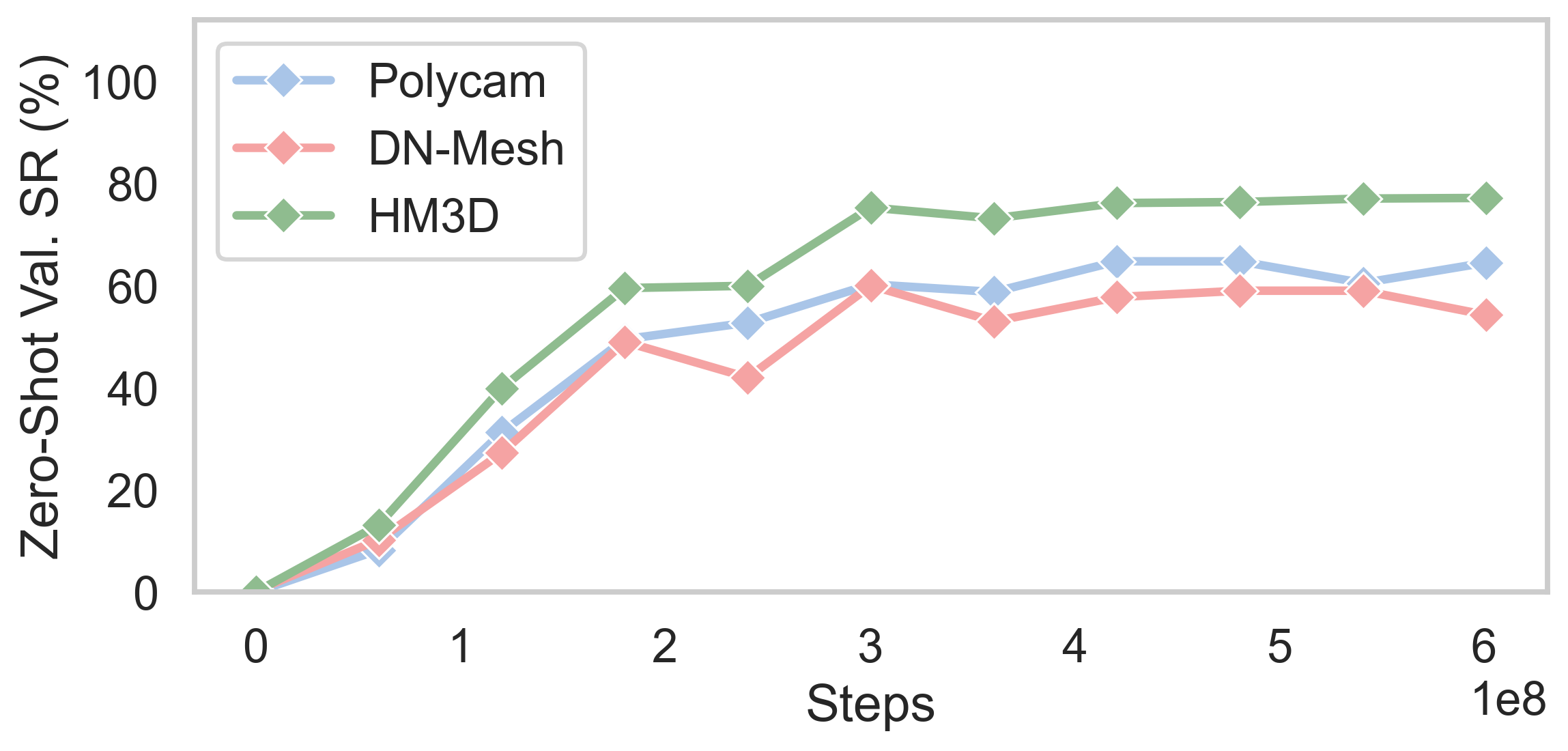}
\caption{\textbf{Average zero-shot success rates on respective validation sets} over different stages of pre-training on HM3D.}
\label{fig:zs_line_chart_hm3d}
\end{figure}

\looseness=-1 \cref{fig:zs_line_chart_hm3d} presents average zero-shot success rates across HM3D validation set, \poly meshes, and \dn meshes at various stages of HM3D policy pre-training. This analysis aims to determine whether continuous performance improvements on HM3D validation scenes translate to improved zero-shot performance on our \cpt scenes. We observe that while performance initially increases, it begins to deteriorate or plateau at approximately 400M steps, despite continued improvements on HM3D validation scenes.

\begin{figure}[t]
\centering
\includegraphics[width=1\linewidth]{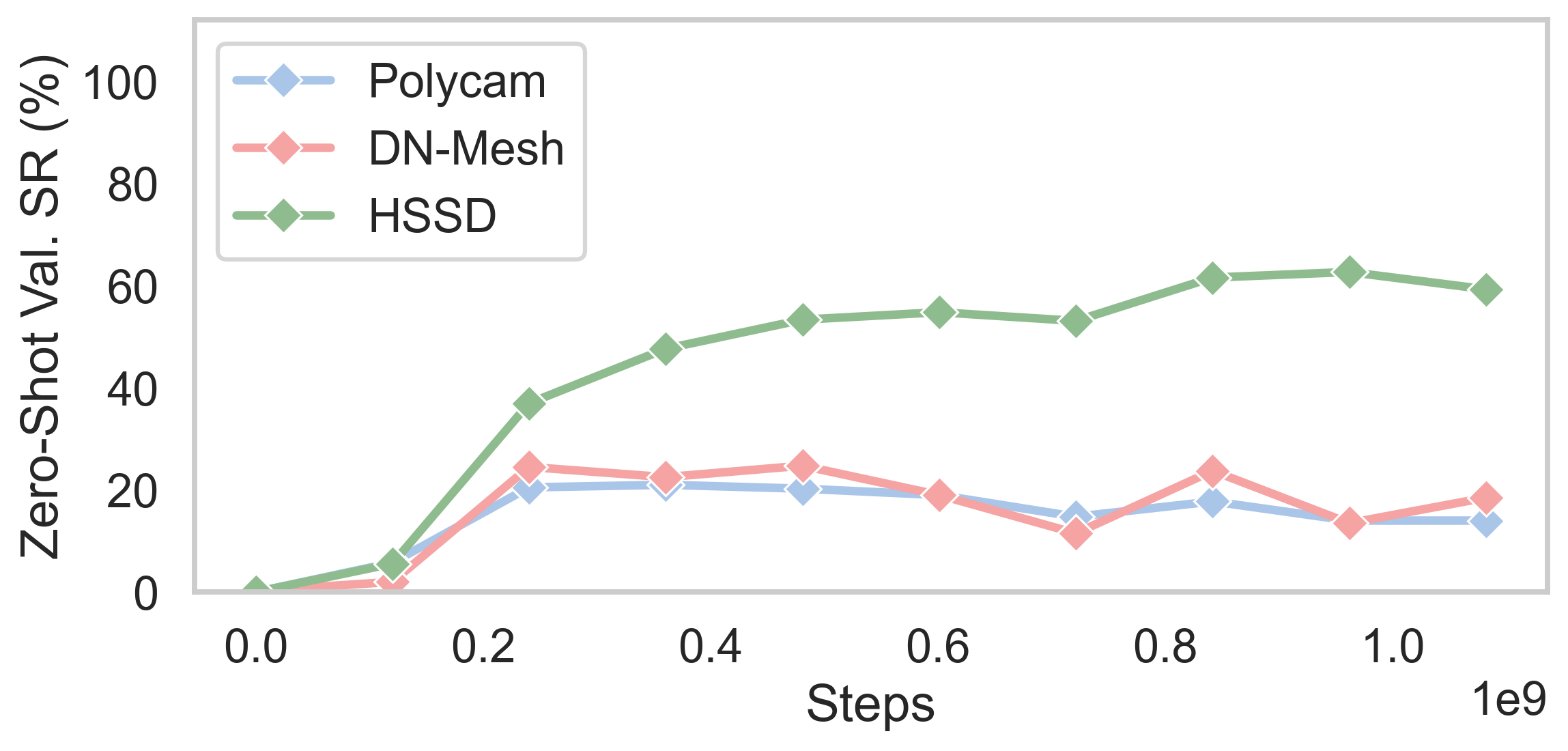}
\caption{\textbf{Average zero-shot success rates on respective validation sets} over different stages of pre-training on HSSD.}
\label{fig:zs_line_chart_hssd}
\end{figure}

\looseness=-1 \cref{fig:zs_line_chart_hssd} shows a similar experiment conducted using the HSSD dataset. Up to 300M steps, improvements in HSSD validation performance correspond to slight improvements in zero-shot performance on our \cpt scenes. However, performance plateaus for \cpt, showing no further gains despite continued improvement on HSSD validation scenes. For HM3D pre-training, \poly meshes outperform \dn meshes in success rates, while for HSSD, the trend reverses slightly, highlighting the impact of pre-training dataset characteristics on zero-shot generalization.
\section{Conclusion}
\label{sec:conclusion}

In this work, we introduced a comprehensive pipeline for bridging the gap between real-world and simulated environments in training embodied agents using 3D Gaussian Splats (GS) and Polycam. By leveraging the \msr dataset and custom iPhone-captured scenes, we demonstrated an efficient and scalable approach to policy personalization, leveraging 3D scene reconstruction from low-effort collects, enabling high-quality training for the ImageNav task. Our pipeline facilitates accessible and replicable scene collection without requiring specialized hardware or significant costs, making it a practical solution for large-scale embodied AI research.
We evaluated overfitted, zero-shot, and fine-tuned policies, showing that fine-tuning pre-trained policies on real-world scene reconstructions improves sim-to-real transfer. We also analyzed the differences between GS-generated \dn meshes and \poly meshes, finding that \poly more closely resemble real-world scenes.

This work lays the foundation for seamlessly integrating real-world scene captures into simulation, expanding the applicability of embodied AI systems. Another promising avenue is integrating GS directly into training, replacing visual observations and goal images to enhance learning efficiency. Furthermore, we aim to extend the use of Gaussian Splats to more complex embodied AI tasks, such as rearrangement and mobile manipulation, broadening its impact across diverse real-world applications.

{
    \small
    \bibliographystyle{ieeenat_fullname}
    \bibliography{main}

\begin{thebibliography}{60}
\providecommand{\natexlab}[1]{#1}
\providecommand{\url}[1]{\texttt{#1}}
\expandafter\ifx\csname urlstyle\endcsname\relax
  \providecommand{\doi}[1]{doi: #1}\else
  \providecommand{\doi}{doi: \begingroup \urlstyle{rm}\Url}\fi

\bibitem[mat()]{matterportCaptureShare}
{C}apture, share, and collaborate the built world in immersive 3{D} --- matterport.com.
\newblock \url{https://matterport.com/}.
\newblock [Accessed 08-03-2025].

\bibitem[Al-Halah et~al.(2022)Al-Halah, Ramakrishnan, and Grauman]{alhalah2022zeroexperiencerequiredplug}
Ziad Al-Halah, Santhosh~K. Ramakrishnan, and Kristen Grauman.
\newblock Zero experience required: Plug \& play modular transfer learning for semantic visual navigation, 2022.

\bibitem[Bae and Davison(2024)]{bae2024dsine}
Gwangbin Bae and Andrew~J. Davison.
\newblock Rethinking inductive biases for surface normal estimation.
\newblock In \emph{IEEE/CVF Conference on Computer Vision and Pattern Recognition (CVPR)}, 2024.

\bibitem[Barron et~al.(2021)Barron, Mildenhall, Tancik, Hedman, Martin-Brualla, and Srinivasan]{barron2021mipnerfmultiscalerepresentationantialiasing}
Jonathan~T. Barron, Ben Mildenhall, Matthew Tancik, Peter Hedman, Ricardo Martin-Brualla, and Pratul~P. Srinivasan.
\newblock Mip-nerf: A multiscale representation for anti-aliasing neural radiance fields, 2021.

\bibitem[Bhat et~al.(2023)Bhat, Birkl, Wofk, Wonka, and Müller]{bhat2023zoedepthzeroshottransfercombining}
Shariq~Farooq Bhat, Reiner Birkl, Diana Wofk, Peter Wonka, and Matthias Müller.
\newblock Zoedepth: Zero-shot transfer by combining relative and metric depth, 2023.

\bibitem[Bono et~al.(2023)Bono, Antsfeld, Chidlovskii, Weinzaepfel, and Wolf]{bono2023endtoendinstanceimagegoalnavigation}
Guillaume Bono, Leonid Antsfeld, Boris Chidlovskii, Philippe Weinzaepfel, and Christian Wolf.
\newblock End-to-end (instance)-image goal navigation through correspondence as an emergent phenomenon, 2023.

\bibitem[Byravan et~al.(2022)Byravan, Humplik, Hasenclever, Brussee, Nori, Haarnoja, Moran, Bohez, Sadeghi, Vujatovic, and Heess]{byravan2022nerf2realsim2realtransfervisionguided}
Arunkumar Byravan, Jan Humplik, Leonard Hasenclever, Arthur Brussee, Francesco Nori, Tuomas Haarnoja, Ben Moran, Steven Bohez, Fereshteh Sadeghi, Bojan Vujatovic, and Nicolas Heess.
\newblock Nerf2real: Sim2real transfer of vision-guided bipedal motion skills using neural radiance fields, 2022.

\bibitem[Chang et~al.(2017)Chang, Dai, Funkhouser, Halber, Nießner, Savva, Song, Zeng, and Zhang]{chang2017matterport3dlearningrgbddata}
Angel Chang, Angela Dai, Thomas Funkhouser, Maciej Halber, Matthias Nießner, Manolis Savva, Shuran Song, Andy Zeng, and Yinda Zhang.
\newblock Matterport3d: Learning from rgb-d data in indoor environments, 2017.

\bibitem[Chaplot et~al.(2020)Chaplot, Gandhi, Gupta, and Salakhutdinov]{chaplot2020objectgoalnavigationusing}
Devendra~Singh Chaplot, Dhiraj Gandhi, Abhinav Gupta, and Ruslan Salakhutdinov.
\newblock Object goal navigation using goal-oriented semantic exploration, 2020.

\bibitem[Chen et~al.(2024{\natexlab{a}})Chen, Xu, Dharmarajan, Irshad, Cheng, Keutzer, Tomizuka, Vuong, and Goldberg]{chen2024roviaug}
Lawrence~Yunliang Chen, Chenfeng Xu, Karthik Dharmarajan, Muhammad~Zubair Irshad, Richard Cheng, Kurt Keutzer, Masayoshi Tomizuka, Quan Vuong, and Ken Goldberg.
\newblock Rovi-aug: Robot and viewpoint augmentation for cross-embodiment robot learning.
\newblock In \emph{Conference on Robot Learning (CoRL)}, Munich, Germany, 2024{\natexlab{a}}.

\bibitem[Chen et~al.(2024{\natexlab{b}})Chen, Shorinwa, Bruno, Yu, Zeng, Nagami, Dames, and Schwager]{chen2024splatnavsaferealtimerobot}
Timothy Chen, Ola Shorinwa, Joseph Bruno, Javier Yu, Weijia Zeng, Keiko Nagami, Philip Dames, and Mac Schwager.
\newblock Splat-nav: Safe real-time robot navigation in gaussian splatting maps, 2024{\natexlab{b}}.

\bibitem[Dai et~al.(2024)Dai, Xu, Xie, Liu, Wang, and Xu]{dai2024highqualitysurfacereconstructionusing}
Pinxuan Dai, Jiamin Xu, Wenxiang Xie, Xinguo Liu, Huamin Wang, and Weiwei Xu.
\newblock High-quality surface reconstruction using gaussian surfels, 2024.

\bibitem[Deitke et~al.(2022)Deitke, Hendrix, Weihs, Farhadi, Ehsani, and Kembhavi]{deitke2022phone2procbringingrobustrobots}
Matt Deitke, Rose Hendrix, Luca Weihs, Ali Farhadi, Kiana Ehsani, and Aniruddha Kembhavi.
\newblock Phone2proc: Bringing robust robots into our chaotic world, 2022.

\bibitem[Eftekhar et~al.(2021)Eftekhar, Sax, Malik, and Zamir]{eftekhar2021omnidata}
Ainaz Eftekhar, Alexander Sax, Jitendra Malik, and Amir Zamir.
\newblock Omnidata: A scalable pipeline for making multi-task mid-level vision datasets from 3d scans.
\newblock In \emph{Proceedings of the IEEE/CVF International Conference on Computer Vision}, pages 10786--10796, 2021.

\bibitem[Guédon and Lepetit(2023)]{guédon2023sugarsurfacealignedgaussiansplatting}
Antoine Guédon and Vincent Lepetit.
\newblock Sugar: Surface-aligned gaussian splatting for efficient 3d mesh reconstruction and high-quality mesh rendering, 2023.

\bibitem[Hu et~al.(2024)Hu, Yin, Zhang, Cai, Long, Chen, Wang, Yu, Shen, and Shen]{hu2024metric3d}
Mu Hu, Wei Yin, Chi Zhang, Zhipeng Cai, Xiaoxiao Long, Hao Chen, Kaixuan Wang, Gang Yu, Chunhua Shen, and Shaojie Shen.
\newblock Metric3d v2: A versatile monocular geometric foundation model for zero-shot metric depth and surface normal estimation.
\newblock 2024.

\bibitem[Irshad et~al.(2021)Irshad, Ma, and Kira]{irshad2021hierarchical}
Muhammad~Zubair Irshad, Chih-Yao Ma, and Zsolt Kira.
\newblock Hierarchical cross-modal agent for robotics vision-and-language navigation.
\newblock In \emph{Proceedings of the IEEE International Conference on Robotics and Automation (ICRA)}, 2021.

\bibitem[Irshad et~al.(2022)Irshad, Chowdhury~Mithun, Seymour, Chiu, Samarasekera, and Kumar]{9956561}
Muhammad~Zubair Irshad, Niluthpol Chowdhury~Mithun, Zachary Seymour, Han-Pang Chiu, Supun Samarasekera, and Rakesh Kumar.
\newblock Semantically-aware spatio-temporal reasoning agent for vision-and-language navigation in continuous environments.
\newblock In \emph{2022 26th International Conference on Pattern Recognition (ICPR)}, pages 4065--4071, 2022.

\bibitem[Irshad et~al.(2023)Irshad, Zakharov, Liu, Guizilini, Kollar, Gaidon, Kira, and Ambrus]{irshad2023neo360}
Muhammad~Zubair Irshad, Sergey Zakharov, Katherine Liu, Vitor Guizilini, Thomas Kollar, Adrien Gaidon, Zsolt Kira, and Rares Ambrus.
\newblock Neo 360: Neural fields for sparse view synthesis of outdoor scenes.
\newblock 2023.

\bibitem[Irshad et~al.(2024)Irshad, Comi, Lin, Heppert, Valada, Ambrus, Kira, and Tremblay]{irshad2024neuralfieldsroboticssurvey}
Muhammad~Zubair Irshad, Mauro Comi, Yen-Chen Lin, Nick Heppert, Abhinav Valada, Rares Ambrus, Zsolt Kira, and Jonathan Tremblay.
\newblock Neural fields in robotics: A survey.
\newblock \emph{arXiv preprint arXiv:2410.20220}, 2024.

\bibitem[Kadian et~al.(2020)Kadian, Truong, Gokaslan, Clegg, Wijmans, Lee, Savva, Chernova, and Batra]{Kadian_2020}
Abhishek Kadian, Joanne Truong, Aaron Gokaslan, Alexander Clegg, Erik Wijmans, Stefan Lee, Manolis Savva, Sonia Chernova, and Dhruv Batra.
\newblock Sim2real predictivity: Does evaluation in simulation predict real-world performance?
\newblock \emph{IEEE Robotics and Automation Letters}, 5\penalty0 (4):\penalty0 6670–6677, 2020.

\bibitem[Kapelyukh et~al.(2024)Kapelyukh, Ren, Alzugaray, and Johns]{kapelyukh2024dream2realzeroshot3dobject}
Ivan Kapelyukh, Yifei Ren, Ignacio Alzugaray, and Edward Johns.
\newblock Dream2real: Zero-shot 3d object rearrangement with vision-language models, 2024.

\bibitem[Kar et~al.(2022)Kar, Yeo, Atanov, and Zamir]{kar20223d}
O{\u{g}}uzhan~Fatih Kar, Teresa Yeo, Andrei Atanov, and Amir Zamir.
\newblock 3d common corruptions and data augmentation.
\newblock In \emph{Proceedings of the IEEE/CVF Conference on Computer Vision and Pattern Recognition}, pages 18963--18974, 2022.

\bibitem[Kerbl et~al.(2023)Kerbl, Kopanas, Leimk{\"u}hler, and Drettakis]{kerbl3Dgaussians}
Bernhard Kerbl, Georgios Kopanas, Thomas Leimk{\"u}hler, and George Drettakis.
\newblock 3d gaussian splatting for real-time radiance field rendering.
\newblock \emph{ACM Transactions on Graphics}, 42\penalty0 (4), 2023.

\bibitem[Khanna et~al.(2023)Khanna, Mao, Jiang, Haresh, Schacklett, Batra, Clegg, Undersander, Chang, and Savva]{khanna2023habitat}
Mukul Khanna, Yongsen Mao, Hanxiao Jiang, Sanjay Haresh, Brennan Schacklett, Dhruv Batra, Alexander Clegg, Eric Undersander, Angel~X Chang, and Manolis Savva.
\newblock Habitat synthetic scenes dataset (hssd-200): An analysis of 3d scene scale and realism tradeoffs for objectgoal navigation.
\newblock \emph{arXiv preprint arXiv:2306.11290}, 2023.

\bibitem[Krantz et~al.(2022)Krantz, Lee, Malik, Batra, and Chaplot]{krantz2022instancespecific}
Jacob Krantz, Stefan Lee, Jitendra Malik, Dhruv Batra, and Devendra~Singh Chaplot.
\newblock Instance-specific image goal navigation: Training embodied agents to find object instances, 2022.

\bibitem[Lei et~al.(2024)Lei, Wang, Zhou, and Li]{lei2024gaussnavgaussiansplattingvisual}
Xiaohan Lei, Min Wang, Wengang Zhou, and Houqiang Li.
\newblock Gaussnav: Gaussian splatting for visual navigation, 2024.

\bibitem[Li and Pathak(2024)]{li2024objectaware}
Yulong Li and Deepak Pathak.
\newblock Object-aware gaussian splatting for robotic manipulation.
\newblock In \emph{ICRA 2024 Workshop on 3D Visual Representations for Robot Manipulation}, 2024.

\bibitem[Loshchilov and Hutter(2019)]{loshchilov2019decoupledweightdecayregularization}
Ilya Loshchilov and Frank Hutter.
\newblock Decoupled weight decay regularization, 2019.

\bibitem[Low et~al.(2025)Low, Adang, Yu, Nagami, and Schwager]{low2025sousvidecookingvisual}
JunEn Low, Maximilian Adang, Javier Yu, Keiko Nagami, and Mac Schwager.
\newblock Sous vide: Cooking visual drone navigation policies in a gaussian splatting vacuum, 2025.

\bibitem[Majumdar et~al.(2023)Majumdar, Aggarwal, Devnani, Hoffman, and Batra]{majumdar2023zsonzeroshotobjectgoalnavigation}
Arjun Majumdar, Gunjan Aggarwal, Bhavika Devnani, Judy Hoffman, and Dhruv Batra.
\newblock Zson: Zero-shot object-goal navigation using multimodal goal embeddings, 2023.

\bibitem[Majumdar et~al.(2024)Majumdar, Yadav, Arnaud, Ma, Chen, Silwal, Jain, Berges, Abbeel, Malik, Batra, Lin, Maksymets, Rajeswaran, and Meier]{majumdar2024searchartificialvisualcortex}
Arjun Majumdar, Karmesh Yadav, Sergio Arnaud, Yecheng~Jason Ma, Claire Chen, Sneha Silwal, Aryan Jain, Vincent-Pierre Berges, Pieter Abbeel, Jitendra Malik, Dhruv Batra, Yixin Lin, Oleksandr Maksymets, Aravind Rajeswaran, and Franziska Meier.
\newblock Where are we in the search for an artificial visual cortex for embodied intelligence?, 2024.

\bibitem[Marza et~al.(2023)Marza, Matignon, Simonin, Batra, Wolf, and Chaplot]{marza2023autonerftrainingimplicitscene}
Pierre Marza, Laetitia Matignon, Olivier Simonin, Dhruv Batra, Christian Wolf, and Devendra~Singh Chaplot.
\newblock Autonerf: Training implicit scene representations with autonomous agents, 2023.

\bibitem[Meng et~al.(2024)Meng, Wu, Yin, and Zhang]{meng2024beingsbayesianembodiedimagegoal}
Wugang Meng, Tianfu Wu, Huan Yin, and Fumin Zhang.
\newblock Beings: Bayesian embodied image-goal navigation with gaussian splatting, 2024.

\bibitem[Mildenhall et~al.(2020)Mildenhall, Srinivasan, Tancik, Barron, Ramamoorthi, and Ng]{mildenhall2020nerfrepresentingscenesneural}
Ben Mildenhall, Pratul~P. Srinivasan, Matthew Tancik, Jonathan~T. Barron, Ravi Ramamoorthi, and Ren Ng.
\newblock Nerf: Representing scenes as neural radiance fields for view synthesis, 2020.

\bibitem[M\"uller et~al.(2022)M\"uller, Evans, Schied, and Keller]{mueller2022instant}
Thomas M\"uller, Alex Evans, Christoph Schied, and Alexander Keller.
\newblock Instant neural graphics primitives with a multiresolution hash encoding.
\newblock \emph{ACM Trans. Graph.}, 41\penalty0 (4):\penalty0 102:1--102:15, 2022.

\bibitem[Piccinelli et~al.(2024)Piccinelli, Yang, Sakaridis, Segu, Li, Van~Gool, and Yu]{piccinelli2024unidepth}
Luigi Piccinelli, Yung-Hsu Yang, Christos Sakaridis, Mattia Segu, Siyuan Li, Luc Van~Gool, and Fisher Yu.
\newblock {U}ni{D}epth: Universal monocular metric depth estimation.
\newblock In \emph{Proceedings of the IEEE/CVF Conference on Computer Vision and Pattern Recognition (CVPR)}, 2024.

\bibitem[Polycam()]{polycam}
Polycam.
\newblock Polycam.
\newblock Accessed: 2024-11-10.

\bibitem[Puig et~al.(2023)Puig, Undersander, Szot, Cote, Yang, Partsey, Desai, Clegg, Hlavac, Min, et~al.]{puig2023habitat}
Xavier Puig, Eric Undersander, Andrew Szot, Mikael~Dallaire Cote, Tsung-Yen Yang, Ruslan Partsey, Ruta Desai, Alexander Clegg, Michal Hlavac, So~Yeon Min, et~al.
\newblock Habitat 3.0: A co-habitat for humans, avatars, and robots.
\newblock In \emph{The Twelfth International Conference on Learning Representations}, 2023.

\bibitem[Qureshi et~al.(2024)Qureshi, Garg, Yandun, Held, Kantor, and Silwal]{qureshi2024splatsimzeroshotsim2realtransfer}
Mohammad~Nomaan Qureshi, Sparsh Garg, Francisco Yandun, David Held, George Kantor, and Abhisesh Silwal.
\newblock Splatsim: Zero-shot sim2real transfer of rgb manipulation policies using gaussian splatting, 2024.

\bibitem[Ramakrishnan et~al.(2021)Ramakrishnan, Gokaslan, Wijmans, Maksymets, Clegg, Turner, Undersander, Galuba, Westbury, Chang, Savva, Zhao, and Batra]{ramakrishnan2021habitatmatterport3ddatasethm3d}
Santhosh~K. Ramakrishnan, Aaron Gokaslan, Erik Wijmans, Oleksandr Maksymets, Alex Clegg, John Turner, Eric Undersander, Wojciech Galuba, Andrew Westbury, Angel~X. Chang, Manolis Savva, Yili Zhao, and Dhruv Batra.
\newblock Habitat-matterport 3d dataset (hm3d): 1000 large-scale 3d environments for embodied ai, 2021.

\bibitem[Ren et~al.(2024)Ren, Wang, Cai, Tuominen, Kannala, and Rahtu]{ren2024mushroommultisensorhybridroom}
Xuqian Ren, Wenjia Wang, Dingding Cai, Tuuli Tuominen, Juho Kannala, and Esa Rahtu.
\newblock Mushroom: Multi-sensor hybrid room dataset for joint 3d reconstruction and novel view synthesis, 2024.

\bibitem[Research(2023)]{home_robot}
Meta~AI Research.
\newblock Home robot.
\newblock \url{https://github.com/facebookresearch/home-robot}, 2023.
\newblock GitHub repository.

\bibitem[Savva et~al.(2019)Savva, Kadian, Maksymets, Zhao, Wijmans, Jain, Straub, Liu, Koltun, Malik, Parikh, and Batra]{savva2019habitatplatformembodiedai}
Manolis Savva, Abhishek Kadian, Oleksandr Maksymets, Yili Zhao, Erik Wijmans, Bhavana Jain, Julian Straub, Jia Liu, Vladlen Koltun, Jitendra Malik, Devi Parikh, and Dhruv Batra.
\newblock Habitat: A platform for embodied ai research, 2019.

\bibitem[Schulman et~al.(2017)Schulman, Wolski, Dhariwal, Radford, and Klimov]{schulman2017proximalpolicyoptimizationalgorithms}
John Schulman, Filip Wolski, Prafulla Dhariwal, Alec Radford, and Oleg Klimov.
\newblock Proximal policy optimization algorithms, 2017.

\bibitem[Silwal et~al.(2024)Silwal, Yadav, Wu, Vakil, Majumdar, Arnaud, Chen, Berges, Batra, Rajeswaran, Kalakrishnan, Meier, and Maksymets]{silwal2024learnlargescalestudypretrained}
Sneha Silwal, Karmesh Yadav, Tingfan Wu, Jay Vakil, Arjun Majumdar, Sergio Arnaud, Claire Chen, Vincent-Pierre Berges, Dhruv Batra, Aravind Rajeswaran, Mrinal Kalakrishnan, Franziska Meier, and Oleksandr Maksymets.
\newblock What do we learn from a large-scale study of pre-trained visual representations in sim and real environments?, 2024.

\bibitem[Szot et~al.(2021)Szot, Clegg, Undersander, Wijmans, Zhao, Turner, Maestre, Mukadam, Chaplot, Maksymets, Gokaslan, Vondrus, Dharur, Meier, Galuba, Chang, Kira, Koltun, Malik, Savva, and Batra]{szot2021habitat}
Andrew Szot, Alex Clegg, Eric Undersander, Erik Wijmans, Yili Zhao, John Turner, Noah Maestre, Mustafa Mukadam, Devendra Chaplot, Oleksandr Maksymets, Aaron Gokaslan, Vladimir Vondrus, Sameer Dharur, Franziska Meier, Wojciech Galuba, Angel Chang, Zsolt Kira, Vladlen Koltun, Jitendra Malik, Manolis Savva, and Dhruv Batra.
\newblock Habitat 2.0: Training home assistants to rearrange their habitat.
\newblock In \emph{Advances in Neural Information Processing Systems (NeurIPS)}, 2021.

\bibitem[Tancik et~al.(2023)Tancik, Weber, Ng, Li, Yi, Kerr, Wang, Kristoffersen, Austin, Salahi, Ahuja, McAllister, and Kanazawa]{nerfstudio}
Matthew Tancik, Ethan Weber, Evonne Ng, Ruilong Li, Brent Yi, Justin Kerr, Terrance Wang, Alexander Kristoffersen, Jake Austin, Kamyar Salahi, Abhik Ahuja, David McAllister, and Angjoo Kanazawa.
\newblock Nerfstudio: A modular framework for neural radiance field development.
\newblock In \emph{ACM SIGGRAPH 2023 Conference Proceedings}, 2023.

\bibitem[Turkulainen et~al.(2024)Turkulainen, Ren, Melekhov, Seiskari, Rahtu, and Kannala]{turkulainen2024dnsplatterdepthnormalpriors}
Matias Turkulainen, Xuqian Ren, Iaroslav Melekhov, Otto Seiskari, Esa Rahtu, and Juho Kannala.
\newblock Dn-splatter: Depth and normal priors for gaussian splatting and meshing, 2024.

\bibitem[Wijmans et~al.(2019)Wijmans, Kadian, Morcos, Lee, Essa, Parikh, Savva, and Batra]{wijmans2019dd}
Erik Wijmans, Abhishek Kadian, Ari Morcos, Stefan Lee, Irfan Essa, Devi Parikh, Manolis Savva, and Dhruv Batra.
\newblock Dd-ppo: Learning near-perfect pointgoal navigators from 2.5 billion frames.
\newblock In \emph{International Conference on Learning Representations}, 2019.

\bibitem[Wolf et~al.(2024)Wolf, Bracha, and Kimmel]{wolf2024gs2meshsurfacereconstructiongaussian}
Yaniv Wolf, Amit Bracha, and Ron Kimmel.
\newblock Gs2mesh: Surface reconstruction from gaussian splatting via novel stereo views, 2024.

\bibitem[Wu et~al.(2024)Wu, Pan, Wu, Wang, Miao, and Wang]{wu2024rlgsbridge3dgaussiansplatting}
Yuxuan Wu, Lei Pan, Wenhua Wu, Guangming Wang, Yanzi Miao, and Hesheng Wang.
\newblock Rl-gsbridge: 3d gaussian splatting based real2sim2real method for robotic manipulation learning, 2024.

\bibitem[Xia et~al.(2018)Xia, Zamir, He, Sax, Malik, and Savarese]{xia2018gibsonenvrealworldperception}
Fei Xia, Amir Zamir, Zhi-Yang He, Alexander Sax, Jitendra Malik, and Silvio Savarese.
\newblock Gibson env: Real-world perception for embodied agents, 2018.

\bibitem[Xia et~al.(2024)Xia, Lin, Ma, and Wang]{xia2024video2gamerealtimeinteractiverealistic}
Hongchi Xia, Zhi-Hao Lin, Wei-Chiu Ma, and Shenlong Wang.
\newblock Video2game: Real-time, interactive, realistic and browser-compatible environment from a single video, 2024.

\bibitem[Yadav et~al.(2022)Yadav, Ramrakhya, Majumdar, Berges, Kuhar, Batra, Baevski, and Maksymets]{yadav2022offlinevisualrepresentationlearning}
Karmesh Yadav, Ram Ramrakhya, Arjun Majumdar, Vincent-Pierre Berges, Sachit Kuhar, Dhruv Batra, Alexei Baevski, and Oleksandr Maksymets.
\newblock Offline visual representation learning for embodied navigation, 2022.

\bibitem[Yadav et~al.(2023)Yadav, Majumdar, Ramrakhya, Yokoyama, Baevski, Kira, Maksymets, and Batra]{yadav2023ovrlv2simplestateofartbaseline}
Karmesh Yadav, Arjun Majumdar, Ram Ramrakhya, Naoki Yokoyama, Alexei Baevski, Zsolt Kira, Oleksandr Maksymets, and Dhruv Batra.
\newblock Ovrl-v2: A simple state-of-art baseline for imagenav and objectnav, 2023.

\bibitem[Yang et~al.(2024)Yang, Kang, Huang, Zhao, Xu, Feng, and Zhao]{yang2024depthv2}
Lihe Yang, Bingyi Kang, Zilong Huang, Zhen Zhao, Xiaogang Xu, Jiashi Feng, and Hengshuang Zhao.
\newblock Depth anything v2, 2024.

\bibitem[Yu et~al.(2024)Yu, Hari, Srinivas, El-Refai, Rashid, Kim, Kerr, Cheng, Irshad, Balakrishna, et~al.]{yu2024language}
Justin Yu, Kush Hari, Kishore Srinivas, Karim El-Refai, Adam Rashid, Chung~Min Kim, Justin Kerr, Richard Cheng, Muhammad~Zubair Irshad, Ashwin Balakrishna, et~al.
\newblock Language-embedded gaussian splats (legs): Incrementally building room-scale representations with a mobile robot.
\newblock In \emph{2024 IEEE/RSJ International Conference on Intelligent Robots and Systems (IROS)}, pages 13326--13332. IEEE, 2024.

\bibitem[Yu et~al.(2025)Yu, Hari, El-Refai, Dalil, Kerr, Kim, Cheng, Irshad, and Goldberg]{yu2025pogs}
Justin Yu, Kush Hari, Karim El-Refai, Arnav Dalil, Justin Kerr, Chung-Min Kim, Richard Cheng, Muhammad~Z. Irshad, and Ken Goldberg.
\newblock Persistent object gaussian splat (pogs) for tracking human and robot manipulation of irregularly shaped objects.
\newblock \emph{ICRA}, 2025.

\bibitem[Zhu et~al.(2024)Zhu, Wang, Kong, and Wang]{zhu20243dgaussiansplattingrobotics}
Siting Zhu, Guangming Wang, Dezhi Kong, and Hesheng Wang.
\newblock 3d gaussian splatting in robotics: A survey, 2024.

\end{thebibliography}
}

\clearpage
\appendix
\renewcommand{\thesection}{\Alph{section}}
\renewcommand{\thesubsection}{\Alph{section}.\arabic{subsection}}

\setcounter{page}{1}
\maketitlesupplementary

\section{Real-world Deployment on Stretch Robot}
\label{sec:appendix_real_world_methodology}
\begin{figure}[htb]
    \centering
    \includegraphics[width=0.9\linewidth]{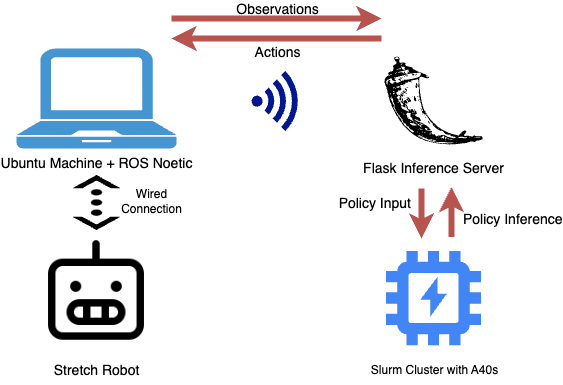}
    \caption{Framework for real-world evaluation on a Stretch robot.}
    \label{fig:real_world_framework}
\end{figure}
Since the stretch robot is not equipped with GPUs, we develop a framework for deploying a policy on a remote cluster, and performing inference using POST requests from the robot. An overview of this arrangement is shown in \cref{fig:real_world_framework}. Specifically, we create a Flask server that runs on a remote compute node, port-forwarded to an Ubuntu 20.04 machine with ROS noetic installed. The same machine is connected to a Stretch robot via ethernet connection. We use the \texttt{home-robot}~\cite{home_robot} repository as framework to convert discretized actions from the policy into continuous space using the ``StretchNavigationClient''. A script running on the machine consumes the ROS topics for image observations from the robot, and passes this along with the goal image to the policy and receives the next action based on these observations.

\begin{figure}
    \centering
    \includegraphics[width=1\linewidth]{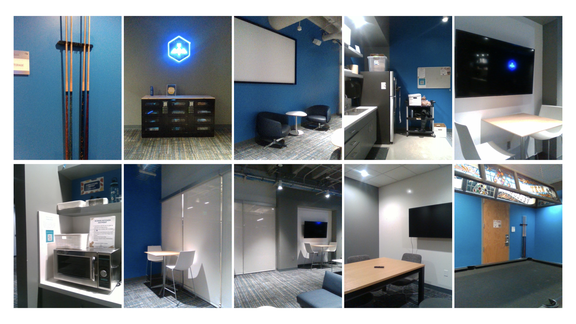}
    \caption{Goal images used in \lng scene}
    \label{fig:goal_images}
\end{figure}

For evaluation in real-world, we use random start-and-goal locations, place the Stretch robot at the goal location, and capture a goal image. Then, we mark these locations using colored masking tape on the floor. The policy is deployed with the goal image passed to the agent during the episode as above, keeping the start location and pose similar for each evaluation. The episode stops when the agent outputs a STOP action or the maximum number of steps (100 in our case) are reached. Finally, the distance of the base is measured from the the goal location using a measuring tape, and the episode is marked successful if the agent reaches within $1m$ of the goal location before stopping. The goal images used are shown in \cref{fig:goal_images}.

\begin{figure}
    \centering
    \includegraphics[width=0.5\linewidth]{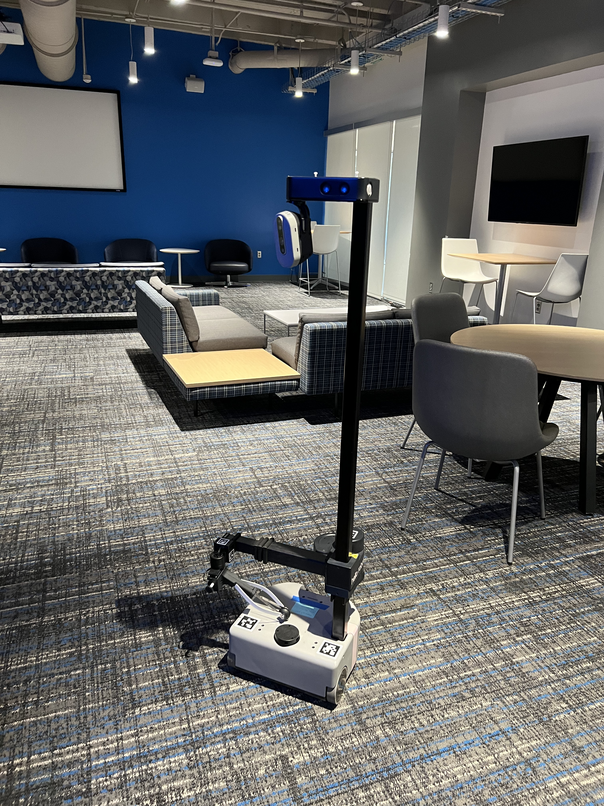}
    \caption{Hello Robot Stretch in \lng scene}
    \label{fig:stretch_robot}
\end{figure}

\section{Generating ImageNav episodes}
\label{subsec:appendix_generate_imagenav_episodes} To generate ImageNav episodes, we build upon Habitat's~\cite{savva2019habitatplatformembodiedai, puig2023habitat} scripts for PointNav generation. During training, the PointNav goal locations are paired with an Image-goal sensor to retrieve the image corresponding to the goal location prior to the start of the episode.

Habitat~\cite{savva2019habitatplatformembodiedai, puig2023habitat} uses navmesh islands to define navigable areas in simulation. For episode generation, we leverage the stretch robot parameters~\cite{silwal2024learnlargescalestudypretrained} to compute the navmeshes.

Valid start and goal locations are sampled during episode generation, following the approach in~\cite{silwal2024learnlargescalestudypretrained}. The validity checks ensure that the goal is reachable, the distance from the start location to the goal is non-zero, and the navmesh island radius exceeds $2m$, thereby avoiding navigation on objects like beds or tables.

\section{Examples for \dn and \poly rendering}

\begin{figure}[ht]
    \centering
    \begin{minipage}{0.49\columnwidth}
        \includegraphics[width=\linewidth]{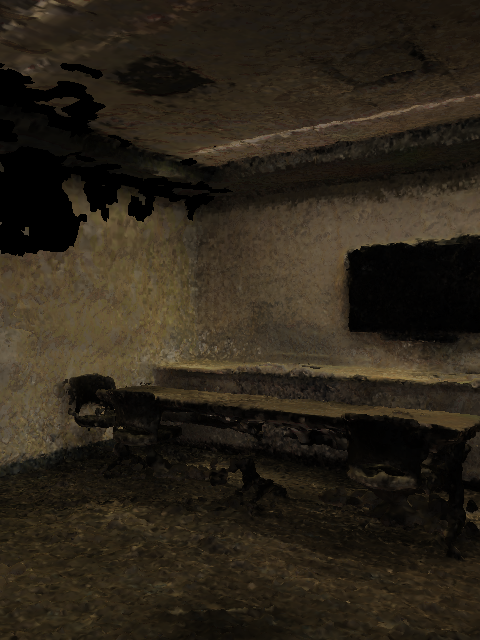}
        \caption{\dn (\cab)}
        \label{fig:dn_rendering}
    \end{minipage}
    \hfill
    \begin{minipage}{0.49\columnwidth}
        \includegraphics[width=\linewidth]{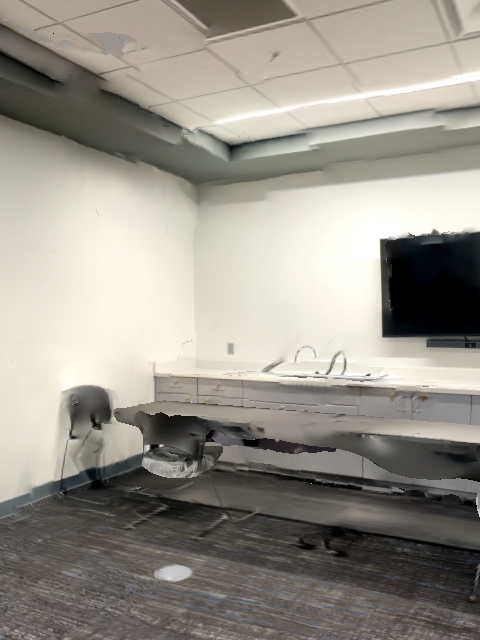}
        \caption{\poly (\cab)}
        \label{fig:polycam_rendering}
    \end{minipage}
    \label{fig:dn_vs_poly}
\end{figure}

\cref{fig:dn_rendering} and \cref{fig:polycam_rendering} shows the rendering differences for the images in Habitat-Sim~\cite{puig2023habitat} for \dn and \poly meshes. \dn meshes have diffused and darker colors, and some holes for regions where capture was not sufficient. In contrast, polycam produces a smoother and more realistic mesh.

\section{Expanded Related Work}
\label{sec: appendix_expanded_related_work}
\subsection{3D Scene and Mesh Reconstruction}
\label{subsec: appendix_related_work_3d_scene}
\looseness=-1 Recent advancements in 3D scene reconstruction have led to the development of several notable approaches. Neural Radiance Fields (NeRF)~\cite{mildenhall2020nerfrepresentingscenesneural} and subsequent methods~\cite{mueller2022instant, irshad2023neo360, barron2021mipnerfmultiscalerepresentationantialiasing} have focused on training and improving neural scene representation techniques. The foundational 3D Gaussian Splatting (GS) method~\cite{kerbl3Dgaussians} has inspired a variety of extensions. DN-Splatter~\cite{turkulainen2024dnsplatterdepthnormalpriors} improves reconstruction quality by incorporating depth and normal regularization with monocular networks. Gaussian Surfels~\cite{dai2024highqualitysurfacereconstructionusing} introduces a technique for flattening 3D Gaussians into 2D surfels, resulting in enhanced mesh reconstruction. GS2Mesh~\cite{wolf2024gs2meshsurfacereconstructiongaussian} incorporates real-world knowledge through stereo-matching to generate smoother meshes. SuGaR~\cite{guédon2023sugarsurfacealignedgaussiansplatting} proposes a fast mesh extraction method using surface-aligned Gaussian splatting.

\subsection{3D Scene Representation in Robotics}
\label{subsec: appendix_related_work_3d_scene_in_robotics}
\looseness=-1 Several studies have explored the use of neural scene representations~\cite{xia2024video2gamerealtimeinteractiverealistic, marza2023autonerftrainingimplicitscene} and 3D Gaussian Splatting (GS) for robotic tasks in simulation. SplatGym~\cite{qureshi2024splatsimzeroshotsim2realtransfer} presents a neural simulator for training robotic control policies using Gaussian Splatting. BEINGS~\cite{meng2024beingsbayesianembodiedimagegoal} leverages 3D Gaussian Splatting as a scene prior to predict future observations and refine navigation strategies. A concurrent work, SOUS-VIDE~\cite{low2025sousvidecookingvisual}, uses a simple drone dynamics model within a high visual fidelity 3DGS reconstruction for training an end-to-end policy and show sim-to-real transfer. We note that our work is a complementary effort focused on image-goal navigation, out-of-domain generalization, the effects of reconstruction quality and training strategies.

\looseness=-1 Recent research has also emphasized the application of neural scene representations for real-world robotic applications~\cite{byravan2022nerf2realsim2realtransfervisionguided, chen2024roviaug, kapelyukh2024dream2realzeroshot3dobject}. Object-Aware Gaussian Splatting~\cite{li2024objectaware} proposes semantic and dynamic 3D representations for robotic manipulation. RL-GSBridge~\cite{wu2024rlgsbridge3dgaussiansplatting} introduces a mesh-based 3D Gaussian Splatting method for zero-shot sim-to-real transfer in manipulation tasks. LEGS~\cite{yu2024language} and POGS~\cite{yu2025pogs} distil semantic features from foundation models into GS to enable downstream navigation and manipulation applications. While these studies investigate Gaussian Splatting for sim-to-real manipulation, our work focuses on using it for personalized sim-to-real Image-Goal Navigation. For additional methods utilizing Neural Fields and Gaussian Splatting in robotics, we refer to the comprehensive surveys~\cite{irshad2024neuralfieldsroboticssurvey, zhu20243dgaussiansplattingrobotics} which examines the applications of Neural fields in robotics.

\section{Mesh Processing using Blender}
\label{sec:appendix_mesh_processing}
\begin{figure}[!htb]
    \centering
    \includegraphics[width=0.9\linewidth]{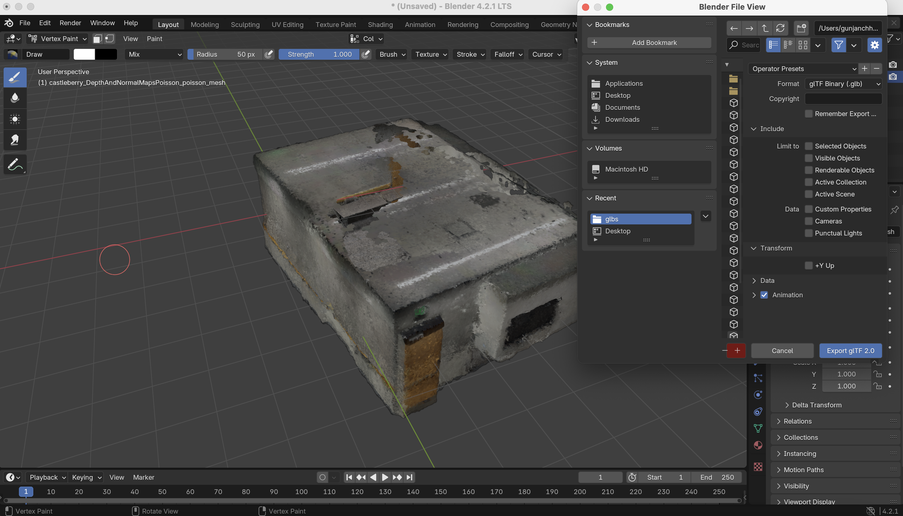}
    \caption{Converting the \texttt{.ply} files generated by DN-Splatter~\cite{turkulainen2024dnsplatterdepthnormalpriors} to \texttt{.glb} files using Blender. Transform \texttt{+Y Up} is kept unchecked following conventions of Habitat-Sim~\cite{puig2023habitat}.}
    \label{fig:blender_processing}
\end{figure}

DN-Splatter~\cite{turkulainen2024dnsplatterdepthnormalpriors} gives us Poisson reconstructed meshes from the Gaussian Splats in \texttt{.ply} format. In order to load these meshes in Habitat-Sim~\cite{puig2023habitat}, we need to convert these to \texttt{.glb} format. In order to do so, we import the \texttt{.ply} in Blender and then export them as \texttt{.glb} files while ensuring that \texttt{+Y Up} transform is unchecked, following the conventions used in Habitat-Sim.

\section{Episode Statistics for Generated Datasets on \cpt Scenes}
\label{sec:appendix_episode_statistics}
\begin{table*}[h!]
\centering
\begin{tabular}{|c|c|c|c|c|c|c|c|}
\hline
\multirow{2}{*}{\textbf{Scene Type}} & \multirow{2}{*}{\textbf{Scene}} & \multicolumn{3}{c|}{\textbf{Train}} & \multicolumn{3}{c|}{\textbf{Val}} \\
\cline{3-8}
 &  & \textbf{Min} & \textbf{Max} & \textbf{Avg.} & \textbf{Min} & \textbf{Max} & \textbf{Avg.} \\
\hline
\multirow{3}{*}{\msr} & \texttt{activity} & 7 & 127 & 59.623 & 17 & 120 & 59.82 \\
 & \texttt{sauna} & 5 & 60 & 34.615 & 7 & 57 & 34.31 \\
 & \texttt{honka} & 6 & 61 & 33.563 & 16 & 63 & 34.16 \\
\hline
\multirow{4}{*}{\cpt} & \lng & 8 & 99 & 51.431 & 18 & 96 & 55.3 \\
 & \csr & 7 & 78 & 35.161 & 11 & 70 & 32.39 \\
 & \cab & 5 & 42 & 20.766 & 5 & 34 & 19.77 \\
 & \pid & 5 & 46 & 19.996 & 6 & 43 & 19.92 \\
\hline
\end{tabular}
\caption{Shortest path lengths for episodes across \msr ~\cite{ren2024mushroommultisensorhybridroom} and \cpt scenes.}
\label{tab:episode_lengths}
\end{table*}

\cref{tab:episode_lengths} presents the shortest path lengths (Min, Max, and Avg.) for both training and validation episodes across the different types of GS-based scenes used in our work. 

Notably, the scale of the \cpt scenes is comparable to that of the \msr ~\cite{ren2024mushroommultisensorhybridroom} scenes, suggesting a similar range of complexity in terms of environment size and structure. In future, we will attempt to collect apartment-scale scenes which require considerably larger number of steps ($\sim$200).

Within the \cpt scenes, there is variation in scale, with certain scenes such as \lng having much larger path lengths compared to others like \cab or \pid, which are conference rooms. 

Both GS training and agent training is likely to be influenced by this scale variability, as the agent must learn to navigate environments that may range from relatively simple, smaller spaces to more expansive, complex ones.

\section{Depth and Normal Encoder Selection for DN-Splatter}
\label{sec:appendix_depth_normals}
Since we do not have ground-truth meshes for our \cpt scenes, we use all 10 \msr ~\cite{ren2024mushroommultisensorhybridroom} scenes as a proxy for evaluation of different depth and normal encoders towards 3D scene reconstruction.

\begin{table*}[h!]
\centering
\begin{tabular}{|c|c|c|c|c|c|c|}
\hline
\textbf{Depth Type} & \textbf{Acc $\downarrow$} & \textbf{Comp $\downarrow$} & \textbf{C-L1 $\downarrow$} & \textbf{NC $\uparrow$} & \textbf{F-score $\uparrow$} \\
\hline
DepthAnything-v2 (Indoor)~\cite{yang2024depthv2} & 0.054 & 0.090 & 0.072 & 0.800 & 0.688 \\
GT & \textbf{0.047} & 0.090 & \textbf{0.068} & \textbf{0.815} & \textbf{0.748} \\
Metric3D-v2~\cite{hu2024metric3d} & 0.052 & 0.089 & 0.071 & 0.804 & 0.700 \\
UniDepth-v2~\cite{piccinelli2024unidepth} & 0.049 & \textbf{0.087} & \textbf{0.068} & 0.813 & 0.719 \\
ZoeDepth~\cite{bhat2023zoedepthzeroshottransfercombining} & 0.061 & 0.091 & 0.076 & 0.780 & 0.621 \\
\hline
\end{tabular}

\caption{Average metrics for \msr Scenes computed against Faro-Scanner ground-truths for different depth types. The normal encoder used is Omnidata~\cite{eftekhar2021omnidata}. Ground truth from iPhone leads to the highest F-scores.}
\label{tab:omnidata_mushroom_metrics}
\end{table*}

\begin{table*}[h!]
\centering
\begin{tabular}{|c|c|c|c|c|c|c|}
\hline
\textbf{Normal Type} & \textbf{Depth Type} & \textbf{Acc $\downarrow$} & \textbf{Comp $\downarrow$} & \textbf{C-L1 $\downarrow$} & \textbf{NC $\uparrow$} & \textbf{F-score $\uparrow$} \\
\hline
DSINE~\cite{bae2024dsine} & GT & 0.046 & 0.090 & 0.068 & 0.815 & 0.750 \\
Metric3D-v2~\cite{hu2024metric3d} & GT & 0.046 & 0.090 & 0.068 & \textbf{0.818} & \textbf{0.752} \\
Omnidata~\cite{eftekhar2021omnidata} & GT & 0.047 & 0.090 & 0.068 & 0.815 & 0.748 \\
\hline
\end{tabular}
\caption{Average metrics for \msr scenes computed against Faro-Scanner ground truths for different normal encoders. The depth used is the ground truth from iPhone. Metric3D-v2 outperforms other normal encoders.}
\label{tab:gt_depth_mushroom_metrics}
\end{table*}

We evaluate various depth encoders using monocular depth predictions and assess their reconstruction performance based on several metrics following \msr ~\cite{ren2024mushroommultisensorhybridroom}: accuracy (Acc), completion (Comp), Chamfer distance (C-L1), normal consistency (NC), and F-score. 
The depth scale for each encoder is defined by the ground truth (GT) from an iPhone, with a transform (scale and bias) learned via gradient optimization to convert the monocular depths from the encoders into the appropriate scale. 

As shown in \cref{tab:omnidata_mushroom_metrics}, the GT depth from iPhone provides the best performance across most metrics, achieving the highest normal consistency (0.815) and F-score (0.748). Among the tested depth encoders, GT consistently outperforms others, confirming its robustness for reconstruction tasks. Since the GT depth encoder performed the best overall, we selected the same for our subsequent evaluations for normal encoders. As shown in \cref{tab:gt_depth_mushroom_metrics}, the Metric3D-v2~\cite{hu2024metric3d} encoder achieves the highest F-score (0.752) when used with GT depth, outperforming other normal encoders - DSINE~\cite{bae2024dsine} and Omnidata~\cite{eftekhar2021omnidata}. Therefore, for 3D reconstruction of our \cpt scenes, we use the GT-depth and Metric3D-v2 normals.
\section{Agent, Training, and Evaluation Details}
\label{sec:appendix_training_details}
\textbf{Agent:} We employ the Hello Robot Stretch robot embodiment as our agent, following the setup in~\cite{silwal2024learnlargescalestudypretrained}. In the Habitat Simulator, the agent is modeled as a cylinder with a height of 1.41m and a radius of 0.3m. The RGB camera sensor is positioned at a height of 1.31m from the ground and is vertically aligned. The sensor outputs images with a resolution of 640$\times$480 (H$\times$W) and a horizontal field of view of 42$^\circ$. The goal sensor is configured with identical parameters to the RGB sensor. During training, the agent's rotation at the goal location is randomly sampled.

\textbf{Training and Evaluation:} We train our agents using DD-PPO~\cite{wijmans2019dd} with ImageNav reward, with each environment generating 64 frames per rollout. The training process includes 2 PPO~\cite{schulman2017proximalpolicyoptimizationalgorithms} epochs, each consisting of 2 mini-batches. Following~\cite{silwal2024learnlargescalestudypretrained}, we use the AdamW optimizer~\cite{loshchilov2019decoupledweightdecayregularization} with a weight decay of $10^{-6}$ and a learning rate of $2.5 \times 10^{-4}$, unless stated otherwise. The visual encoder is kept unfrozen during training and data augmentation is applied. The visual encoder learning rate is set to $1.5 \times 10^{-6}$ for HM3D and other settings, and $1.5 \times 10^{-5}$ for HSSD.  The goal and observation visual encoders share the same weights. The number of environments per GPU is set to 10 for HM3D/HSSD, and 8 for \cpt scenes due to computational constraints. Policies are trained to convergence across all training setups and datasets.
Checkpoints are saved approximately every $\sim3M$ steps and evaluated on the validation sets of the corresponding datasets. The best checkpoint is selected based on the highest success rate (SR) achieved on the validation set.
For training, we utilized 16 NVIDIA A40 GPUs per policy per dataset. For simulation-based evaluations, we utilize a single A40 GPU with one environment for \cpt scenes and \msr scenes, and 20 environments for HM3D/HSSD. For real-world evaluations, we employ a single environment and a single A40 GPU. For details on real-world setup, please refer to \cref{sec:appendix_real_world_methodology}.

\textbf{Visual Encoder:} The observation and goal images are resized to 160$\times$120 before being input into the \texttt{VC-1-Base} visual encoder~\cite{majumdar2024searchartificialvisualcortex}. Patch embeddings are processed through a ``compression layer''~\cite{yadav2023ovrlv2simplestateofartbaseline}, which comprises of a 2D convolution followed by group normalization, to generate the final embeddings.

\textbf{Policy:} The policy is implemented as a 2-layer LSTM that takes as input the previous action embedding, visual observation embeddings, and goal image embeddings following \cite{silwal2024learnlargescalestudypretrained}. It outputs one of the following discrete actions: \texttt{MOVE\_FORWARD}, \texttt{TURN\_LEFT}, \texttt{TURN\_RIGHT}, or \texttt{STOP}.

\textbf{Reward Function:} The reward function is adapted from ~\cite{silwal2024learnlargescalestudypretrained, yadav2023ovrlv2simplestateofartbaseline}, using the same hyperparameters as in~\cite{silwal2024learnlargescalestudypretrained}: success weight $c_s = 5.0$, angle success weight $c_a = 5.0$, goal radius $r_g = 1.0$, angle threshold $\theta_g = 25^\circ$, and slack penalty $\lambda = 0.01$. The collision penalty is defined as $c_{coll} = 0.03$. 

The full reward function is detailed in \cref{eq:imagenav_reward}.

\begin{align}
r_t &= c_s \times \left( \left[ d_t < r_g \right] \land \left[ a_t = \text{STOP} \right] \right) \notag \\
    &+ c_a \times \left( \left[ \theta_t < \theta_g \right] \land \left[ a_t = \text{STOP} \right] \right) \notag \\
    &+ \left( \hat{\theta}_{t-1} - \hat{\theta}_t \right) \notag \\
    &+ \left( d_{t-1} - d_t \right) - \gamma \notag \\
    &- c_{\text{coll}} \times \left[ \text{collision = true} \right]
\label{eq:imagenav_reward}
\end{align}
\section{Real-world metrics for successful episodes}
\label{sec:appendix_real_world_metrics}

\cref{tab:real_world_metrics} shows the distance-to-goal (in centimeters) and the number of steps taken on average for the successful episodes for each type of policy we deploy in the real world. We note that some of these average values may not include enough samples to be statistically significant. Therefore, we refrain from making conclusions about which policy is more efficient based on these, and provide the values for book-keeping purposes.
\begin{table}[h]
\centering
\begin{tabular}{|l|r|r|}
\hline
\textbf{Policy} & \textbf{D2G (cm)} & \textbf{\#Steps} \\
\hline
HM3D            & 44.20                                & 45.60                                \\
HM3D-FT-Poly    & 48.29                                & 58.14                                \\
HM3D-FT-DN      & 29.64                                & 52.86                                \\
HSSD            & 98.50                                & 38.00                                \\
HSSD-FT-Poly    & 37.10                                & 54.00                                \\
HSSD-FT-DN      & 40.87                                & 36.25                                \\
\hline  
\end{tabular}
\caption{Average distance to goal and number of steps in the real-world \lng scene for successful episodes for each policy.}
\label{tab:real_world_metrics}
\end{table}
\section{Simulation results on additional real-world scenes}
\label{sec:additional_results}

In \cref{tab:additional_scene_results}, we present additional results for HM3D zero-shot and fine-tuned policies across several scenes. Five of these are from the \msr dataset—\texttt{koivu}, \texttt{classrm2}, \texttt{kokko}, \texttt{coffeerm}, and \texttt{vr\_rm}. We also include one real-world scene, \texttt{conf\_c}, that we captured using our methodology. Fine-tuning consistently improves performance across all mesh types. The improvements are modest in smaller scenes (e.g., \texttt{koivu}, \texttt{coffeerm}), where the zero-shot policy already performs well. In contrast, larger scenes (e.g., \texttt{conf\_c}, \texttt{vr\_rm}) benefit significantly from fine-tuning.

\begin{table}[h]
\centering
\begin{tabular}{|l|l|r|r|}
\hline
Scene & Mesh & HM3D-ZS & HM3D-FT \\
\hline
\texttt{koivu}         & \dn   & 0.87 & 0.98 \\
\texttt{classrm2}     & \dn   & 0.61 & 0.97 \\
\texttt{kokko}         & \dn   & 0.66 & 0.99 \\
\texttt{coffeerm}  & \dn   & 0.90 & 1.00 \\
\texttt{vr\_rm}      & \dn   & 0.39 & 0.99 \\
\hline
\texttt{conf\_c}       & \dn   & 0.16 & 0.84 \\
                       & Poly & 0.50 & 0.98 \\
\hline
\end{tabular}
\caption{Validation success rates on additional scenes from \msr and \cpt datasets.}
\label{tab:additional_scene_results}
\end{table}

\begin{figure}[h]
\centering
\includegraphics[width=1\linewidth]{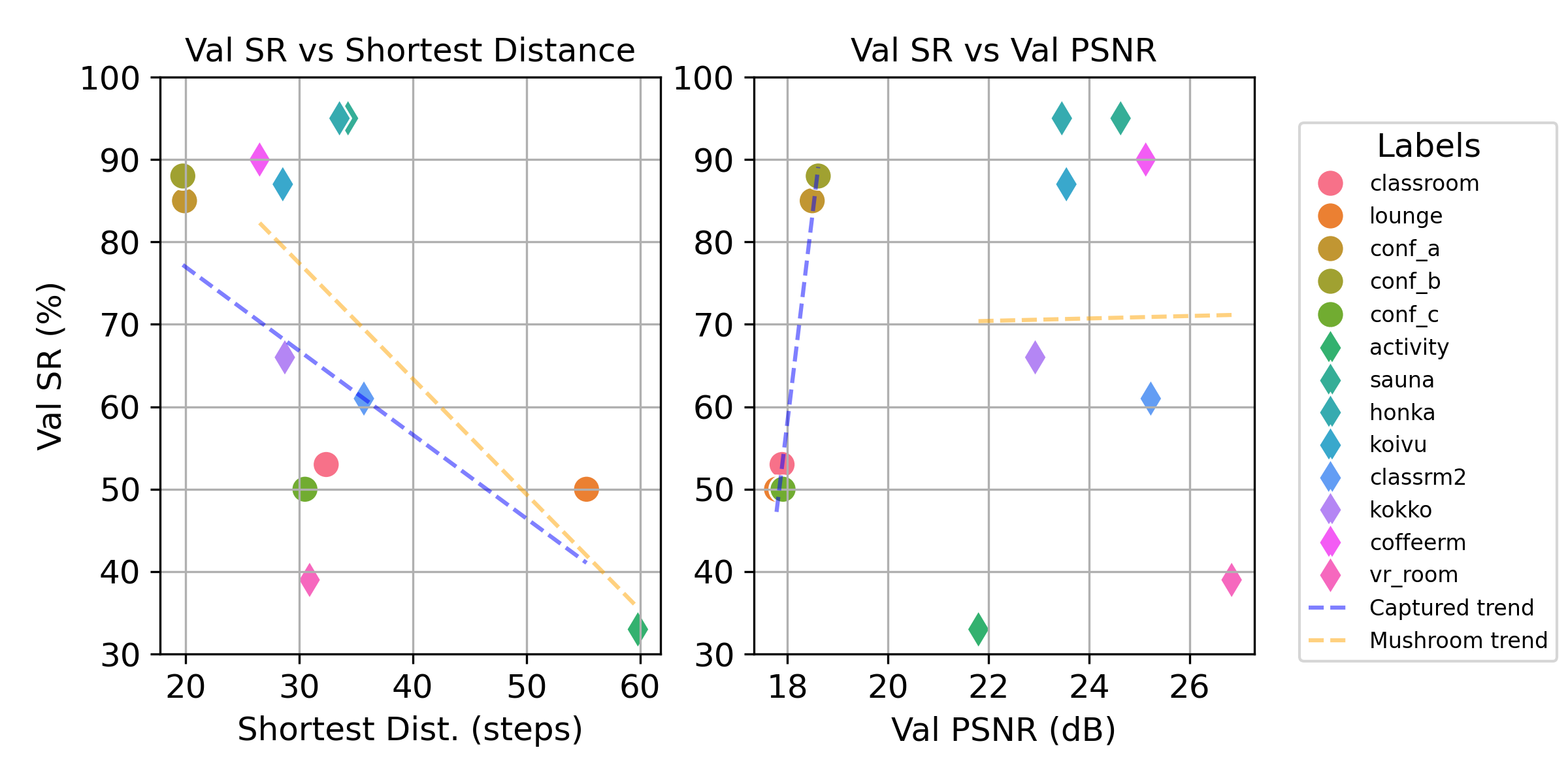}
\caption{\textbf{HM3D zero-shot validation success rates on \dn meshes vs. validation GS PSNRs vs. average shortest distances of validation episodes.} The zero-shot SR is inversely correlated with scene scale and directly correlated with GS validation PSNR. \texttt{vr\_rm} is an outlier in the \msr dataset and requires further analysis.}
\label{fig:val_sr_vs_psnr_vs_shortest_dist_additional_scenes}
\end{figure}

We also visualize the relationship between HM3D zero-shot success rate, validation PSNR, and average shortest path distance in \cref{fig:val_sr_vs_psnr_vs_shortest_dist_additional_scenes}. We observe a consistent trend with the previous results: the zero-shot success rate decreases as the size of the scene increases, and improves as the validation PSNR of the trained Gaussian Splat (GS) increases. 

An exception is \texttt{vr\_rm}, which appears as an outlier in the \msr dataset and warrants further investigation. Aside from this case, the remaining scenes from \msr generally show improved zero-shot performance as PSNR increases.

\section{Failure modes for HM3D zero-shot on \lng scene in simulation}
\label{sec:failure_modes}

To assess the role of visual fidelity, we analyze zero-shot HM3D policy trajectories and identified failure modes in the \lng scene using both \dn and \poly meshes (\cref{tab:termination_reasons_transposed}) in simulation. We found significant increase in ``Maximum Steps Reached'' failures with the \dn mesh, indicating low visual fidelity prevents the policy from matching goal images— a key limitation of the pre-trained policy.

\begin{table}[h]
\centering
\begin{tabular}{|l|r|r|}
\hline
\textbf{Termination Reason} & \textbf{\poly} & \textbf{\dn} \\
\hline
Target Reached (success)          & 74 & 55 \\
Early Stop (Goal Not Visible)     & 3  & 5  \\
Early Stop (Goal Visible)         & 11 & 16 \\
Early Stop (Similar Goal Visible) & 4  & 5  \\
Maximum Steps Reached             & 8  & 19 \\
\hline
\end{tabular}
\caption{Sim failure analysis for HM3D-ZS on \lng.}
\label{tab:termination_reasons_transposed}
\end{table}

\section{Overfitting vs. Fine-Tuning}
\label{sec:appendix_overfit_vs_finetune}

In this section, we clarify why the overfitting results shown in \cref{fig:overfit_combined} may appear stronger than the fine-tuning results in \cref{fig:fine-tune-hm3d}. Fine-tuning requires significantly fewer steps because the pre-trained policy already captures general navigation behaviors. We also deliberately limit the number of fine-tuning steps to preserve the policy’s ability to generalize to real-world environments—leveraging knowledge acquired during large-scale pre-training across diverse scenes.

This pre-training not only enables better generalization in the real world (see \cref{fig:real_world_results}) but also substantially reduces overall training time, which is critical for rapid deployment.

In contrast, our overfitting experiments involve training policies exclusively on simulated versions of specific real scenes. These policies typically achieve 80--90\% success within 20--30M steps, but their performance is limited to the training environment. For completeness, we early stop around 100M steps, although most gains saturate earlier.

\cref{tab:overfit_20m} reports the best validation success rates of overfitted policies at 20M steps. Despite strong performance in simulation, these policies fail to generalize to real-world settings due to factors such as lighting variations, sensor noise, and actuation inaccuracies.

\begin{table}[h]
\centering
\begin{tabular}{|l|r|r|}
\hline
Scene & DN val. SR. & Poly val. SR. \\
\hline
\lng             & 0.74 & 0.74 \\
\texttt{classrm}             & 1.00 & 0.85 \\
\pid             & 0.99 & 1.00 \\
\cab             & 1.00 & 1.00 \\
\texttt{honka}   & 1.00 & ---  \\
\texttt{sauna}   & 0.99 & ---  \\
\texttt{activity}& 0.79 & ---  \\
\hline
\end{tabular}
\caption{Overfitted validation success rates after 20M steps.}
\label{tab:overfit_20m}
\end{table}

\end{document}